\documentclass[twoside,11pt]{article}

\usepackage{blindtext}

\usepackage{subcaption}

\usepackage[preprint]{jmlr2e}

\usepackage{tcolorbox}
\tcbuselibrary{listings, skins, breakable, xparse}
\usepackage{tabularx}
\usepackage{booktabs}

\usepackage{algorithm}
\usepackage{algorithmic}
\usepackage{amsmath}

\newcommand{\method}{VLM-PBRS }
\newcommand{\methodnospace}{VLM-PBRS}
\newtcolorbox{prompttextbox}[1][]{
  left=0pt,
  right=0pt,
  top=0pt,
  bottom=0pt,
  colback=gray!10,
  colframe=black!70,
  boxrule=0.5pt,
  #1
}
\tcbset{
    listing engine=listings,
    colback=gray!10, 
    colframe=black!70, 
    listing only,
    hbox,
    left=1mm,
    right=1mm,
    top=1mm,
    bottom=1mm,
    boxsep=5pt,
    arc=0pt,
    outer arc=0pt,
    boxrule=0.5pt,
    enhanced jigsaw, 
}

\usepackage{lastpage}
\jmlrheading{23}{2026}{1-\pageref{LastPage}}{1/26; Revised 1/26}{1/26}{21-0000}{Henrik M\"uller and Daniel Kudenko}

\ShortHeadings{Automating Potential-based Reward Shaping with Vision Language Model Guidance}{M\"uller and Kudenko}
\firstpageno{1}

\begin{document}

\title{Automating Potential-based Reward Shaping with Vision Language Model Guidance}

\author{\name Henrik M\"uller \email hmueller@l3s.de \\
       \addr L3S Research Center\\
       Leibniz University Hannover\\
       Hannover, 30167, Germany
       \AND
       \name Daniel Kudenko \email kudenko@l3s.de \\
       \addr L3S Research Center\\
       Leibniz University Hannover\\
       Hannover, 30167, Germany}

\editor{My editor}

\maketitle

\begin{abstract}
Sparse rewards are inherently challenging for reinforcement learning agents as they lack intermediate feedback to guide exploration and to correctly attribute the sparse success rewards to relevant parts of the trajectory. Naive reward shaping can induce reward hacking, yielding policies that exploit auxiliary signals instead of solving the intended task. Potential‑based reward shaping (PBRS) guarantees preservation of the optimal policy set, but requires the definition of a heuristic potential function over the state space.
In this work, we introduce the VLM‑guided PBRS framework \method that learns the potential function directly from vision language model (VLM) feedback. We query a lightweight VLM to obtain preferences over image pairs and train a model of the potential function using these preferences. As this approach is based on potential-based reward shaping, it preserves the original optimal policies, and removes the need for expert‑designed reward shaping terms.
Because large VLMs are prohibitively expensive to invoke repeatedly during policy learning, we employ smaller, more computationally efficient VLMs. Although the resulting preference labels are less accurate, empirical evidence shows that the preference labels can still be used to accelerate learning.
We validate our method empirically in the Meta-World and Franka Kitchen environments and highlight the connection between VLM preference label accuracy and sample efficiency improvements.
Our contributions are threefold: (1) the first application of VLM preference‑based learning to synthesize a potential function for PBRS, (2) a principled, low‑cost solution that leverages small VLMs, and (3) extensive empirical demonstration of improved sample efficiency and robustness to reward hacking.
\end{abstract}

\begin{keywords}
  reinforcement learning, potential-based reward shaping, VLM preference-based learning
\end{keywords}

\section{Introduction}
Reward is the single objective that drives all reinforcement learning (RL) agents, yet designing a reward function that both captures the intended behavior and can be learned efficiently remains a challenge, which can be domain‑specific and labor‑intensive. 
In principle, a sparse reward that simply signals task completion could suffice in many cases and an agent would learn to achieve the task by maximizing its expected return. 
In practice, however, sparse rewards are challenging to optimize as most trajectories have the same return and therefore cannot guide exploration before reaching the goal for the first time by chance.
As a result, RL agents tend to require many samples to discover a successful strategy given a sparse reward. 
This problem is commonly alleviated through reward shaping, wherein auxiliary signals are added to the base reward to guide exploration and accelerate learning~\citep{hu2020utilize, memarian2021selfsupervisedrs, trott2019keepingyourdistance, devidze2022explorationrs}. 
But, arbitrary shaping can give rise to reward hacking, wherein maximizing the expected reward no longer corresponds to the original objective~\citep{randlov1998bicycle}. 

Potential‑based reward shaping (PBRS) offers a theoretically sound alternative: by adding the difference of a potential function between successive states, PBRS preserves the optimal policy set while supplying richer learning signals. The challenge, however, is that PBRS still demands a carefully engineered potential function that captures the true structure of the task. Crafting such a potential function typically requires expert knowledge and substantial engineering effort. 

Recent years have seen an explosion of vision language models (VLMs) capable of reasoning about both images and natural language~\citep{radford2021learning, sontakke2023roboclip, lu2024ovis}. 
In RL-VLM-F~\citep{wang2024rlvlmf}, human input for reward design was replaced with VLMs, demonstrating that VLM‑based reward learning can reduce the burden of hand‑crafted reward design and can generalize across domains. 
But the correctness of the learned reward function depends on accurate VLM output, which leads to an inherent risk of reward hacking.
Building on that line of research, we propose a novel VLM‑guided reward shaping framework called \method that learns from VLM-generated preference labels over images to automatically construct a PBRS function.

A key practical consideration is the computational cost of invoking large VLMs during RL training. State‑of‑the‑art models are prohibitively expensive for repeated query cycles, which are necessary as new parts of the state space are explored.
Due to the inherent policy invariance of PBRS, \method does not share the strict label accuracy requirement of RL-VLM-F.
As a result, we are able to use small, less expensive VLMs that still retain a significant fraction of the vision‑language reasoning capabilities and, in contrast to RL-VLM-F, are able to avoid additional LLM calls to generate the preference labels. 
Although the resulting preference labels are potentially less accurate, 
in potential-based reward shaping, this will only reduce the improvement in sample efficiency, rather than risking to worsen the learned policy~\citep{mueller2025incorrectincomplete}.

In summary, our contributions are:
\begin{itemize}
    \item We introduce \methodnospace, which learns a potential function for PBRS based on VLM preference-based learning, requiring only a textual description of the goal and the visual observations of the environment.
    \item We establish that the policy invariance guarantee of PBRS relaxes the label accuracy requirements of direct reward modeling, enabling the use of smaller, computationally efficient VLMs without compromising policy optimality.
    \item We verify the efficacy of our proposed method empirically in the Meta-World and Franka Kitchen environments and analyze the connection between label accuracy and sample efficiency improvements.
\end{itemize}

\begin{figure*}[tb]
\centering
\includegraphics[width=0.7\textwidth]{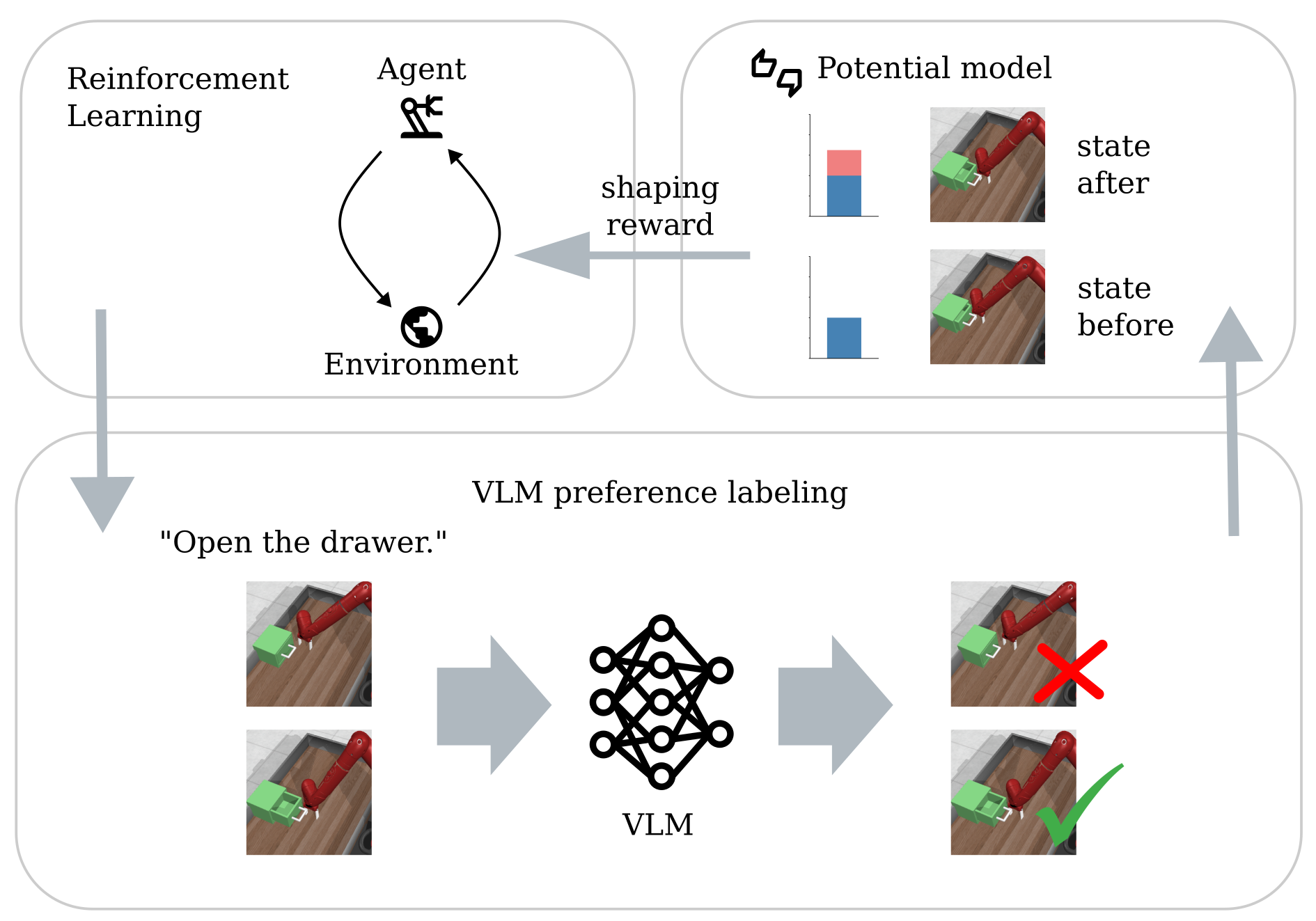}
\caption{Overview of VLM-PBRS.}
\label{fig:overview}
\end{figure*}


\section{Related Work}
We organize the related work around two themes central to this paper: foundational approaches to reward shaping from domain knowledge, and the emerging literature on learning reward functions from scratch using foundation models.

\subsection{Reward Shaping from Domain Knowledge}
Potential-based reward shaping (PBRS) is commonly used to incorporate prior knowledge of the task to guide the agent in solving difficult tasks. To do so, the prior knowledge is encoded in the so-called potential function, which is a heuristic for the \textit{goodness} of a state. Previous works have built on different knowledge representations to automate the definition of potential functions.
They created the potential function from automata specifying a sequence of goals to achieve~\citep{Hasanbeig2021DeepSynthAS}, from user provided linear temporal logic formulas~\citep{elbarbari2022tlrl}, or directly from demonstrations in imitation learning settings~\citep{wang2023dshape, brys2015demonstration, suay2016demonstration, wu2021demonstrations}. 

A first approach has started to use large language models (LLMs) to generate domain knowledge to encode in a potential function~\citep{zhang2025llmbackground}. But, they rely on a hand-crafted translation of the environment state and actions into text to ground the LLM interactions. This 
adds a significant amount of manual labor when adapting the method to new environments, and producing meaningful translations of low-level actions and states can be a challenge in itself.

In contrast to these works, our approach requires far less human input and design. Rather than designing the abstract domain knowledge and having to ground it in the environment observations, or directly demonstrating the desired behaviors, the human only has to input a short high-level goal description for our method.

\subsection{Reward Learning using Foundation Models}
Building on vision language models, like CLIP~\citep{radford2021learning} and RoboCLIP~\citep{sontakke2023roboclip}, there have been works on automatically creating rewards for image observations using a textual goal description~\citep{chan2023visionlanguage,cui2022zeroshot, mahmoudieh2022zero, ma2023liv, sontakke2023roboclip, adeniji2023language, rocamonde2023visionlanguage}. These works define the reward through alignment measures between image embeddings and the language goal embedding in a learned latent space. The resulting reward functions tend to be very noisy and are often poorly aligned with the intended reward~\citep{sontakke2023roboclip, mahmoudieh2022zero}.
Accordingly, some works did task-specific fine-tuning~\citep{ma2023liv, mahmoudieh2022zero}, which we can avoid by using larger, more general VLMs in a more noise robust framework.

Recent works in learning reward functions from scratch using foundation models have found success in querying for preference labels over states and learning the reward model from these labels.

In RL-VLM-F~\citep{wang2024rlvlmf}, reward functions are learned given a natural language description of the goal by querying VLMs for preferences over pairs of environment observations, which are then used to learn a proxy reward model that can be used by an RL agent to learn a policy achieving the goal. We modify RL-VLM-F to provide guidance towards a goal through reward shaping instead of using it to define the policy and goal. 
We reuse the reward model as a potential function for potential-based reward shaping, which additionally softens the requirements on the correctness of the reward model and therefore the correctness of the labels provided by the VLM. 

\citet{lin-2024-navigating-noisy} use LLM preference labels to learn a reward model. Notably, their approach is based on score differences similar to our use in potential-based reward shaping.
But as they differ from the potential-based reward shaping framework their method can learn sub-optimal policies. 
Due to their use of LLMs, they also require a hand-crafted translation of the observations into natural language.

A comparison of different approaches using LLMs as reward sources can be found in \citet{klissarov2025modelingcapabilities}. 
Their study contrasts several reward‑generation strategies: learning reward models from AI preferences, embedding‑based reward signals (created through a similarity metric over the state and goal embeddings), reward as code approaches, and directly querying an LLM for either scalar rewards or actions. The authors find that embedding‑based rewards do not align well with the ground truth task objectives, and that reward‑as‑code, although better aligned, is limited by requiring a useful symbolic representation of the environment. Direct scalar reward outputs, while straightforward, fail to capture the nuances required in more complex environments. Finally, they observe that LLMs exhibit unreliable zero‑shot reasoning about environmental dynamics, which undermines the model's effectiveness as generic reward providers. 
In contrast, the most effective approach was learning reward models from AI feedback, though still facing problems with learning noisy reward functions and unintentionally incentivizing unwanted behaviors.

Previous approaches on using foundation models to create intrinsic reward functions to shape an otherwise uninformative reward have been based on LLMs~\citep{klissarov2024motif, chu2023accelerating}. They created their shaping reward based on preference labels over pairs of textual descriptions of environment states. In contrast to our work, they have to create a manual grounding function to translate the low-level states into text, which can be non-trivial as the exact ground-truth state information can be difficult to capture and similarly be challenging to accurately describe using language. This challenge is inherently solved by the innate grounding capabilities of VLMs. Moreover, these works are not based on potential-based reward shaping and can therefore (in contrast to our approach) bias the policy that is being learned.



\section{Preliminaries}

\subsection{Reinforcement Learning}
A Markov Decision Process (MDP) $M$ is a tuple $M = (S, A, T, \gamma, R)$, where $S$ is the set of states, $A$ is the set of actions, $T: S \times A \to S$ is the transition function defining the new state after executing an action in the current state, $\gamma \in (0, 1]$ is the discount factor, and $R$ is the reward function assigning a scalar feedback to any transition in the environment. The goal of reinforcement learning is to learn a policy (of which action to choose depending on the current state) to maximize the discounted sum of returns: $\sum_t \gamma^t r_t$.

We focus in this work on goal-directed environments with sparse rewards. In this case, the reward function only assigns a reward of one for transitions into a small set of goal states. Otherwise, for all other transitions, the reward is zero. Due to the lack of dense reward feedback, all unsuccessful trajectories have the same sum of rewards. Therefore, the RL agent has no intermediate information to guide its search before reaching any goal state. Moreover, once a trajectory is successful, correctly attributing the success to the relevant parts of the trajectory adds another challenge.

\subsection{Potential-based Reward Shaping}
A potential function $\Phi(s)$ is a heuristic of \textit{goodness} of a state. For the sparse, goal-directed MDPs we are focusing on, it is used to evaluate how close a state is to the goal.
Given a potential function $\Phi(s)$, the potential-based shaping function $F$ is defined as:
\begin{equation}    
\label{eq:pbrs-shaping}
    F(s, a, s') =  \gamma \Phi(s') - \Phi(s)
\end{equation}

The shaping function is then added to the original reward to get the shaped reward function $R'$:
\begin{equation}    
\label{eq:shaped-reward}
    R'(s, a, s') = R(s, a, s') + F(s, a, s')
\end{equation}
The potential-based reward function can then be used by an RL agent to guide its intermediate exploration by trying to maximize the potential function.
The MDP $M'= (S, A, T, \gamma, R')$ with the shaped reward function has the same set of optimal policies as the original, unshaped MDP M~\citep{ng1999invariance}. Independent of the exact potential function (and its correctness), the shaped MDP $M'$ will have the same ordering of policies and as such also the same set of optimal policies.
The only requirement on the potential function $\Phi(s)$ in MDPs with finite episode lengths is that the potential of all terminal states must be zero in order to guarantee policy invariance~\citep{grzes2017episodicPBRS}.

Notably, even if the potential function dynamically changes over time, the theoretical guarantee of policy invariance still holds~\citep{devlin2012dynamicPBRS}, which allows us to continuously improve our potential function over the entire training duration without introducing a bias towards suboptimal policies.

\subsection{Preference-based Reward Learning}
\label{preference-based}
Preference-based reward learning~\citep{wirth2017prefrl, christiano2017humanpref} uses human preference annotations of policy examples to automatically learn a reward function that when optimized by reinforcement learning learns the desired behavior. 
Following the RL-VLM-F~\citep{wang2024rlvlmf} approach, we use VLMs as a scalable proxy for the human annotators to generate preference labels in this work.

Formally, the input is a pair of segments $(\sigma^i, \sigma^j)$. The annotator then assigns a preference to the pair $\sigma^i \succ \sigma^j$, where $\succ$ indicates that segment $\sigma^i$ is preferred over segment $\sigma^j$. For this work, the segments have a length of one and are single images. A preference label $y$ is assigned according to:
\begin{equation*}    
y = 
    \begin{cases}
      0 & \text{, if } \sigma^i \succ \sigma^j\\
      1 & \text{, if } \sigma^i \prec \sigma^j\\
      -1 & \text{, no preference}
    \end{cases}
\end{equation*}

The probability of preferring one segment over the other is based on the Bradley-Terry model~\citep{bradley1952rank} and defined using a parameterized preference function $r_\psi$:
\begin{equation*}    
P_\psi [\sigma^i \succ \sigma^j ] = \frac{\exp (r_\psi(\sigma^i))}{\exp(r_\psi(\sigma^i)) + \exp(r_\psi(\sigma^j))}
\end{equation*}

The parameterized preference function $r_\psi$ is then commonly optimized via binary cross-entropy loss:
\begin{equation}
\label{eq:celoss}
\mathcal{L}_{\text{CE}} = - \mathbb{E}_{(\sigma_1, \sigma_0, y)\sim D} [ \mathbb{I}(\sigma_0 \succ \sigma_1)\log P_\psi [\sigma_0 \succ \sigma_1] + \mathbb{I}(\sigma_1 \succ \sigma_0)\log P_\psi [\sigma_1 \succ \sigma_0] ]
\end{equation}

Alternatively, it has been shown that the mean absolute error (MAE) loss is better suited for noisy labels~\citep{ghosh2017robustloss}, which has also been applied in the context of reward learning from VLM preference labels~\citep{luu2025ratingbased}. We experiment with the MAE loss in Section~\ref{sec:ablation-loss}, but find no improvement in performance. In this case, the loss function is defined by:

\begin{equation}    
\label{eq:maeloss}
\mathcal{L}_{\text{MAE}} = - \mathbb{E}_{(\sigma_1, \sigma_0, y)\sim D} \frac{1}{2} \Bigl[ \bigl | (|y - 1|) - P_\psi [\sigma_0 \succ \sigma_1] \bigr | + | y - P_\psi [\sigma_1 \succ \sigma_0] | \Bigr]
\end{equation}



\begin{figure*}[tb]
\centering
\includegraphics[width=\textwidth]{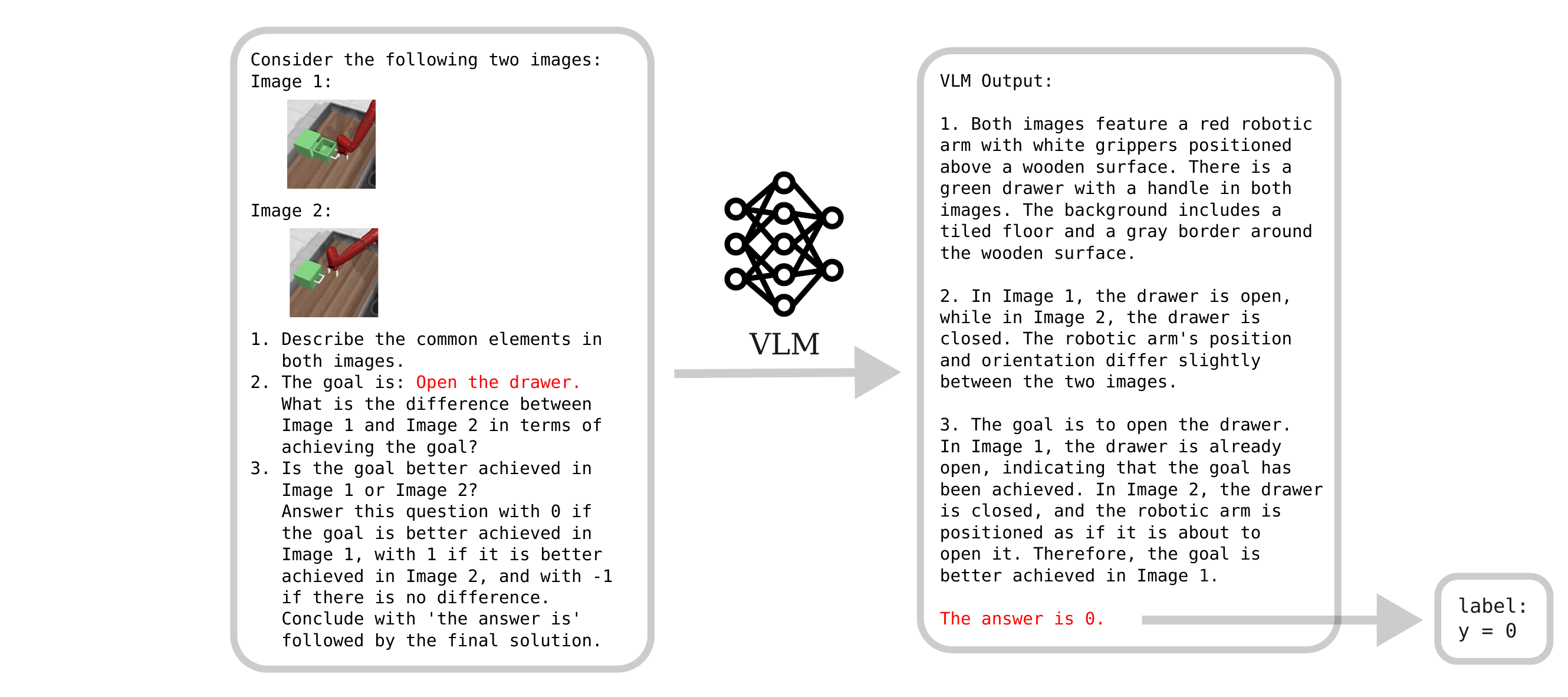}
\caption{Prompt pipeline for an example image pair from the drawer-open task of the Meta-World environment. The images and goal description are added to the prompt template on the left. The filled template is input to the VLM. The final label is then extracted from the VLM output on the right.}
\label{fig:prompt-example}
\end{figure*}

\begin{algorithm}[tb]
\caption{\method} \label{alg:overview}
\begin{algorithmic}[1]
\REQUIRE text description of goal $l$
\REQUIRE environment interaction steps $T$, policy gradient update steps $\mathcal{N}_{\pi}$, potential update steps $\mathcal{N}_{r}$, VLM labeling frequency $K$, number of preference queries per labeling batch $M$

\STATE Initialize policy $\pi_\theta$ and preference model $r_\psi$
\STATE Initialize the preference buffer $\mathcal{D} \leftarrow \emptyset$, replay buffer $\mathcal{B} \leftarrow \emptyset$, image observation buffer $\mathcal{I} \leftarrow \emptyset$

\STATE {{\textsc{// Initial exploration}}}
\FOR{each exploration iteration}
    \STATE $\pi_\theta, \mathcal{B}, \mathcal{I} \leftarrow$ Reinforcement learning loop in Algorithm~\ref{alg:rl-loop}
\ENDFOR

\FOR{each iteration}
    \STATE {{\textsc{// VLM labeling and preference model learning}}}
    \IF{iteration \% $K == 0$}
    \FOR{$m = 1$ to $M$}
    \STATE Randomly sample two images $(\sigma^0, \sigma^1)$ from buffer $\mathcal{I}$
    \STATE{Query VLM with $(\sigma^0, \sigma^1)$ and task goal $l$ for label $y$}
    \STATE Store preference $\mathcal{D} \leftarrow \mathcal{D}\cup \{(\sigma^0, \sigma^1, y)\}$
    \ENDFOR
    \FOR{$n=1$ to $\mathcal{N}_{r}$}
    \STATE Sample minibatch $\{(\sigma^0, \sigma^1,y)_j\}_{j=1}^D\sim\mathcal{D}$
    \STATE Optimize $r_\psi$ via loss in Equation~\eqref{eq:celoss}
    \ENDFOR
    \STATE Recompute shaping rewards in replay buffer $\mathcal{B}$ using updated $r_\psi$
    \ENDIF

    \STATE {{\textsc{// Policy improvement and data collection}}}
    \STATE $\pi_\theta, \mathcal{B}, \mathcal{I} \leftarrow$ Reinforcement learning loop in Algorithm~\ref{alg:rl-loop}
\ENDFOR
\end{algorithmic}
\end{algorithm}

\begin{algorithm}[tb]
\caption{Reinforcement Learning Loop} \label{alg:rl-loop}
\begin{algorithmic}[1]
\REQUIRE policy $\pi_\theta$ and preference model $r_\psi$, replay buffer $\mathcal{B}$, image observation buffer $\mathcal{I}$
\REQUIRE environment interaction steps $T$, policy gradient update steps $\mathcal{N}_{\pi}$

\FOR{$t=1$ to $T$}
    \STATE  Collect state $s_{t+1}$, image $I_{t+1}$, reward $r_{t+1}$ by taking $a_t \sim \pi_\theta(a_t | s_t)$
    \STATE Compute next potential $\Phi(s_{t+1}) \leftarrow r_\psi (I_{t+1})$
    \STATE Compute shaped reward $r'_{t+1}$ according to Eq.~\eqref{eq:pbrs-shaping} and \eqref{eq:shaped-reward}
    \STATE Add transition $\mathcal{B} \leftarrow \mathcal{B} \cup \{(s_t,a_t,s_{t+1},r'_{t+1})\}$
    \STATE Add image observation $\mathcal{I} \leftarrow \mathcal{I} \cup \{I_{t+1}\}$
\ENDFOR 
\FOR{$n=1$ to $\mathcal{N}_{\pi}$}
    \STATE Sample minibatch $\{(s_t, a_t, s_{t+1}, r'_{t+1})_j\}_{j=1}^B\sim\mathcal{B}$
    \STATE Optimize policy $\pi_\theta$ using minibatch with any RL algorithm
\ENDFOR
\RETURN $\pi_\theta, \mathcal{B}, \mathcal{I}$
\end{algorithmic}
\end{algorithm}

\section{Method}
\label{method}

In \methodnospace, we follow the general structure used in reward learning from preference labels. We iterate between querying for new preference labels, learning the preference proxy model, and improving the policy via reinforcement learning to collect new data~\citep{christiano2017humanpref, lee2021pebble, wang2024rlvlmf}.
An overview of our method can be found in Figure~\ref{fig:overview}.

During an initial unguided exploration phase (without VLM preference labeling or preference model learning) the initial sets of trajectories and environment images are collected.
After the unguided exploration phase, the main potential learning and policy improvement loop begins. 

First, we sample pairs of images uniform at random from the set of images collected during the previous environment interactions. These image pairs together with the goal description of the task (in natural language) are then added to the prompt. 
The prompt template is task agnostic with the only task grounding being provided through the task description. 
The prompt is structured to first ground the output vocabulary through common elements in the image pair, then describing the differences between the images with respect to the goal, and lastly decide on the label in a machine-readable output ("the answer is \textit{y}").
The filled prompt is then fed into the VLM, the label $y$ is extracted from the output text, and the labeled image pair is added to the preference training set. 

An example of the VLM labeling pipeline can be found in Figure~\ref{fig:prompt-example}. 
The prompt template can be found in the Appendix in Figure~\ref{fig:prompt-ours} for the Meta-World experiments and in Figure~\ref{fig:prompt-frankakitchen} for the Franka Kitchen experiments. To achieve more accurate labels in the visually more challenging Franka Kitchen environment, we extended the prompt with a system prompt describing the task to improve the label accuracy adapted from \citet{zhao2025consistent}. In contrast to RL-VLM-F, we simply extract the label $y$ from a single VLM call rather than using an additional LLM call to generate it from the VLM output. This enables us to improve computational efficiency at the cost of potentially lower labeling accuracy.

The training of the preference model $r_\psi$ is continued whenever a batch of new preference labels is available. The model is trained following the cross entropy loss defined in Eq.~\eqref{eq:celoss} (or respectively for the MAE loss in Eq.~\eqref{eq:maeloss}). 

Following the considerations of \citet{mueller2025effectivepbrs}, the potential function should be bounded and its maximum should be smaller than the goal reward to ensure that the shaped goal reward remains positive, which creates an incentive to reach and terminate in the goal state to enable the RL agent to properly exploit the original goal rewards.
To this end, we add a sigmoid activation function ($\sigma$) and a scaling factor $\lambda$ to transform the preference model into the potential function. 
The potential function is therefore defined as follows:
\begin{eqnarray}
\label{eq:potential-def}
\Phi(s) = 
    \begin{cases}
      0 & \text{, if } s \text{ is a terminal state}\\
      \lambda \sigma (r_\psi(s)) & \text{, otherwise }
    \end{cases}
\end{eqnarray}
In our experiments, the reward for reaching a goal state is always one, so we set $\lambda = 0.9$.

Once the potential function is (re-)trained, the shaped rewards in the replay buffer are recomputed to make use of the most recent, most informed version of the potential function. The algorithm then continues with standard reinforcement learning using the new shaping rewards to guide its exploration, to improve its policy, and to collect new data. Thus, our method trades additional VLM computations for a reduction in required environment interactions.

\subsection{Reward Shaping Reduces Label Accuracy Requirements}

The central advantage of our approach over the dense reward modeling in RL-VLM-F lies in relaxing the accuracy requirements on the learned preference model.
In RL-VLM-F, the learned preference model $r_\psi$ serves directly as the reward function and must therefore closely approximate the true task. We instead use $r_\psi$ as the potential function for potential-based reward shaping. Since PBRS preserves the set of optimal policies for any potential function, the optimality of the learned policies does not hinge on $r_\psi$ perfectly capturing the ground truth reward.
Even incomplete or partially incorrect shaping functions can nonetheless improve the convergence speed of the agent~\citep{mueller2025incorrectincomplete}. \method can therefore leverage smaller, computationally efficient VLMs whose label accuracy would be insufficient for direct reward modeling.

This relaxed accuracy requirement motivates a corresponding simplification of the labeling pipeline. 
RL-VLM-F uses a two-stage process: a VLM first describes and compares two observed images, and a second query to an LLM (or the same VLM) derives the final preference label from that output.
We instead query the VLM in a single prompt to provide the common elements in both images, the difference between the images regarding the goal description, and the output label in the form "The answer is \textit{y}", where $y\in {-1, 0, 1}$ is the final label as described in Section~\ref{preference-based}. 
An example of the full prompt pipeline can be found in Figure~\ref{fig:prompt-example}.


\section{Experiment Setup}
In this section, we first describe the environments used in our experiments, then the methods we compared, and finally the open-weight vision language models we used.

\subsection{Environments}

\begin{figure*}[tb]
\centering
\begin{subfigure}[b]{0.2\textwidth}
    \centering
    \includegraphics[width=\textwidth]{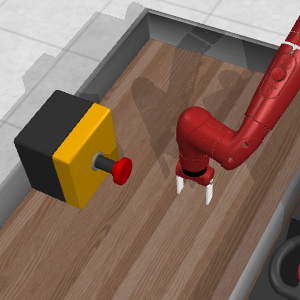}
    \caption{button-press}
\end{subfigure}
\begin{subfigure}[b]{0.2\textwidth}
    \centering
    \includegraphics[width=\textwidth]{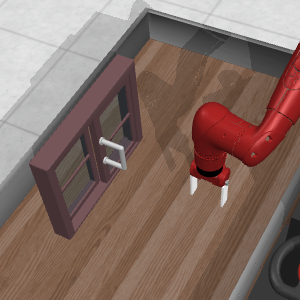}
    \caption{window-open}
\end{subfigure} \\
[1.em]
\begin{subfigure}[b]{0.2\textwidth}
    \centering
    \includegraphics[width=\textwidth]{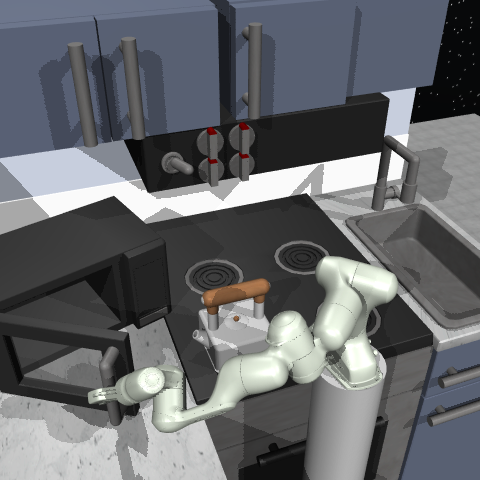}
    \caption{microwave}
\end{subfigure}
\begin{subfigure}[b]{0.2\textwidth}
    \centering
    \includegraphics[width=\textwidth]{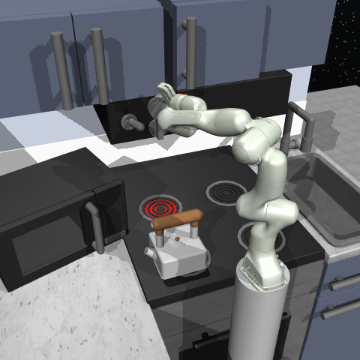}
    \caption{top-burner}
\end{subfigure}
\begin{subfigure}[b]{0.2\textwidth}
    \centering
    \includegraphics[width=\textwidth]{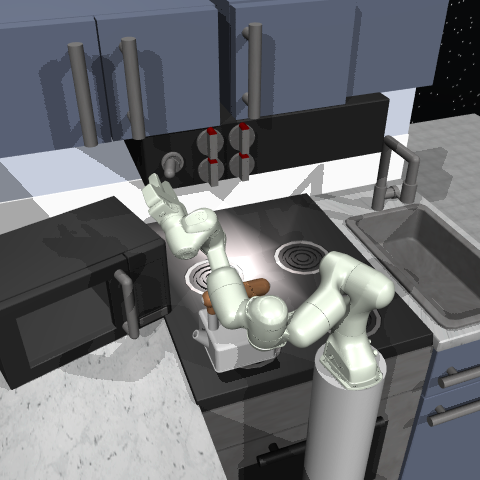}
    \caption{light-switch}
\end{subfigure} \\

\caption{Example observations from Meta-World (top) and Franka Kitchen (bottom).}
\label{fig:examples-envs}
\end{figure*}

In \textit{Meta-World}~\citep{yu2020metaworld}, the agent controls a robot arm and has to interact with the object on the table in front of it to solve a number of different tasks. The 4-dimensional action specifies the 3-dimensional displacement of the end-effector and one value to adjust the gripper. Each observation is 18-dimensional, and describes the position of the robot and the target objects. We include experiments on the following four tasks: \textit{door-open}, \textit{window-open}, \textit{drawer-open}, and \textit{button-press}. At the beginning of each episode, the position of the object to interact with is randomized.

To evaluate how well our method generalizes to a more realistic and visually challenging setting,
we adopt the \textit{Franka Kitchen}~\citep{gupta2019relay, fu2020d4rl} environment.
In \textit{Franka Kitchen}, the agent controls a 9-DoF Franka robot arm placed in a simulated kitchen environment containing numerous distractor objects. Each task is to interact with one of the objects in the environment and move it into its goal position. All other objects serve as visual noise for the VLM preference labeling. Each observation is represented as a 59-dimensional vector, encapsulating the joint state of the robot and all objects in the kitchen environment. We experiment on the following tasks: \textit{microwave}, \textit{light-switch}, and \textit{top-burner}.

We use the original goal descriptions for most tasks, which can be found in the Appendix in Table~\ref{tab:task_goal_description}. In \textit{Franka Kitchen}, we adapt the task descriptions for two tasks to clearly define the goal for the VLM. We extend the description for \textit{light-switch} to specify the unusual location of the light switch on the panel above the stove, and corrected the descriptions for \textit{top-burner} to specify the top \textit{left} burner.
We slightly modify the visual rendering for \textit{Meta-World}, which is only used for the VLM querying approaches. We remove the green and red orbs that originally highlighted the ends of the gripper and the goal position in order to mitigate potential artificial distractions. For all tasks within each environment, we use the same default camera view. 
Every episode consists of 500 time steps for Meta-World and 280 time steps for Franka Kitchen. All tasks have the same reward structure with a reward of one per step once the target object has reached its goal position and rewards of zero for all other transitions. Episodes only end after the set number of steps to encourage policies able to reach and remain in the goal position.

\subsection{Baselines}
We use SAC~\citep{haarnoja2018sac} as the underlying RL algorithm for all baselines. The hyperparameters are based on the choices from RL-VLM-F~\citep{wang2024rlvlmf}. An extensive overview of the hyperparameters can be found in Tables~\ref{table:sac-hyperparameters}, \ref{table:pbrs-hyperparameters}, and~\ref{table:vlm-hyperparameters} in the Appendix. In our experiments, we compare the following approaches:

\begin{itemize}
    \item \textit{Sparse}: The original goal-directed sparse reward. Reward of one when moving into goal states, and reward of zero for all other transitions.

    \item \textit{Dense}: Human-designed dense rewards specifically designed to accelerate the sample efficiency of RL algorithms in these environments with the potential risk of leading to a sub-optimal learned policy. These reward functions include additional terms based on privileged task information (otherwise not available to the agent) to directly incentivize moving the robot gripper to the object or respectively its handle, and moving the object towards the goal position. These reward functions were provided in the \textit{Meta-World} environment. In the \textit{Franka Kitchen} environment, we defined the dense reward as the sum of the negative distances between gripper and object handle, and object to target position.

    \item \textit{RL-VLM-F}: Reward learning from VLM feedback~\citep{wang2024rlvlmf}. Intended to learn a reward model given a task description from VLM preference labels over image pairs when no external environment reward function is available. For a fair comparison, we add the sparse reward signal to the reward model output. The main difference to our approach is in the direct use of the reward model rather than using potential-based reward shaping.

    \item \textit{\method}: Our method for automatically constructing a potential-based reward shaping from VLM preference labels as described in Section~\ref{method}.
\end{itemize}

\subsection{VLM}
We use two open-weight VLMs: Ovis2~\citep{lu2024ovis} for the Meta-World experiments and Qwen3-VL~\citep{yang2025qwen3} for the Franka Kitchen experiments. To improve computational efficiency, we use the 16B parameter version of Ovis2 and the 8B parameter version of Qwen3-VL. Both can be run locally on a single 40GB A100 GPU. 
They are the smallest versions that still reliably followed the prompt template and produced the label output statement ("the answer is \textit{y}.") correctly, which enables us to avoid having to query an additional LLM on the VLM output to generate the preference labels (as was done in \textit{RL-VLM-F}~\citep{wang2024rlvlmf}).


\begin{figure*}[tb]
\centering
\begin{subfigure}[b]{0.35\textwidth}
    \centering
        \includegraphics[width=\textwidth]{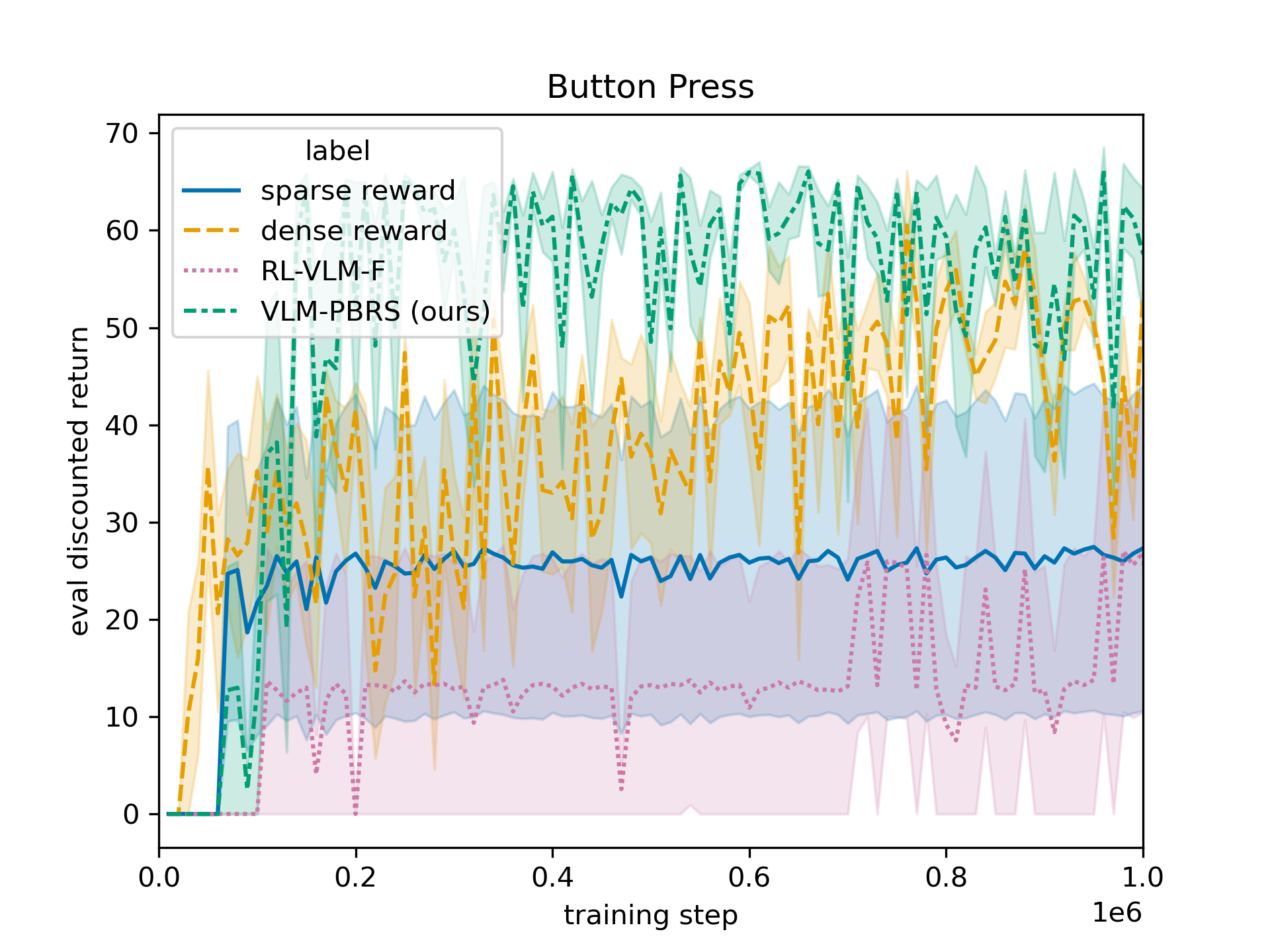}
    \caption{button-press}
\end{subfigure}
\begin{subfigure}[b]{0.35\textwidth}
    \centering
        \includegraphics[width=\textwidth]{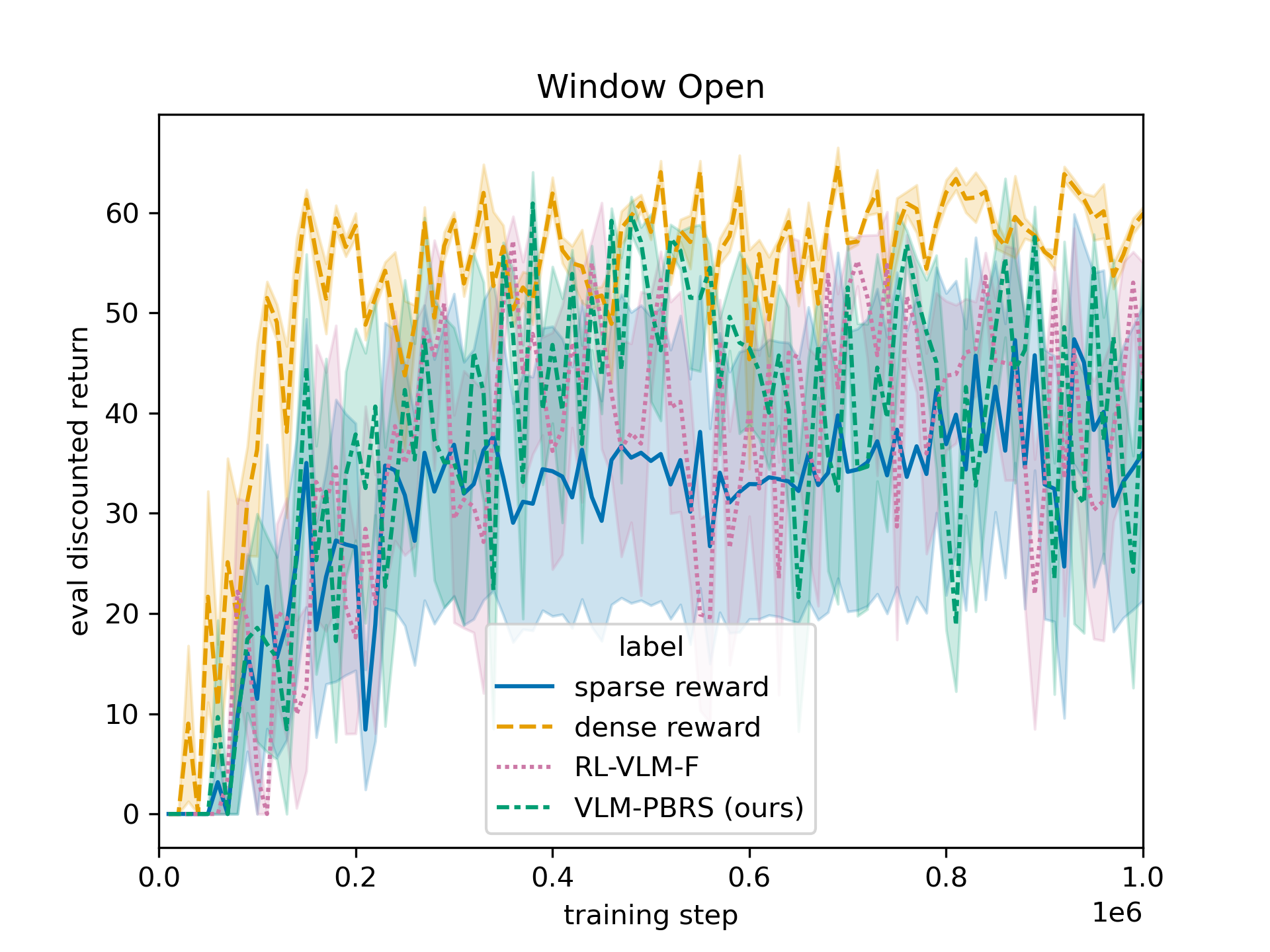}
    \caption{window-open}
\end{subfigure} \\

\begin{subfigure}[b]{0.35\textwidth}
    \centering
    \includegraphics[width=\textwidth]{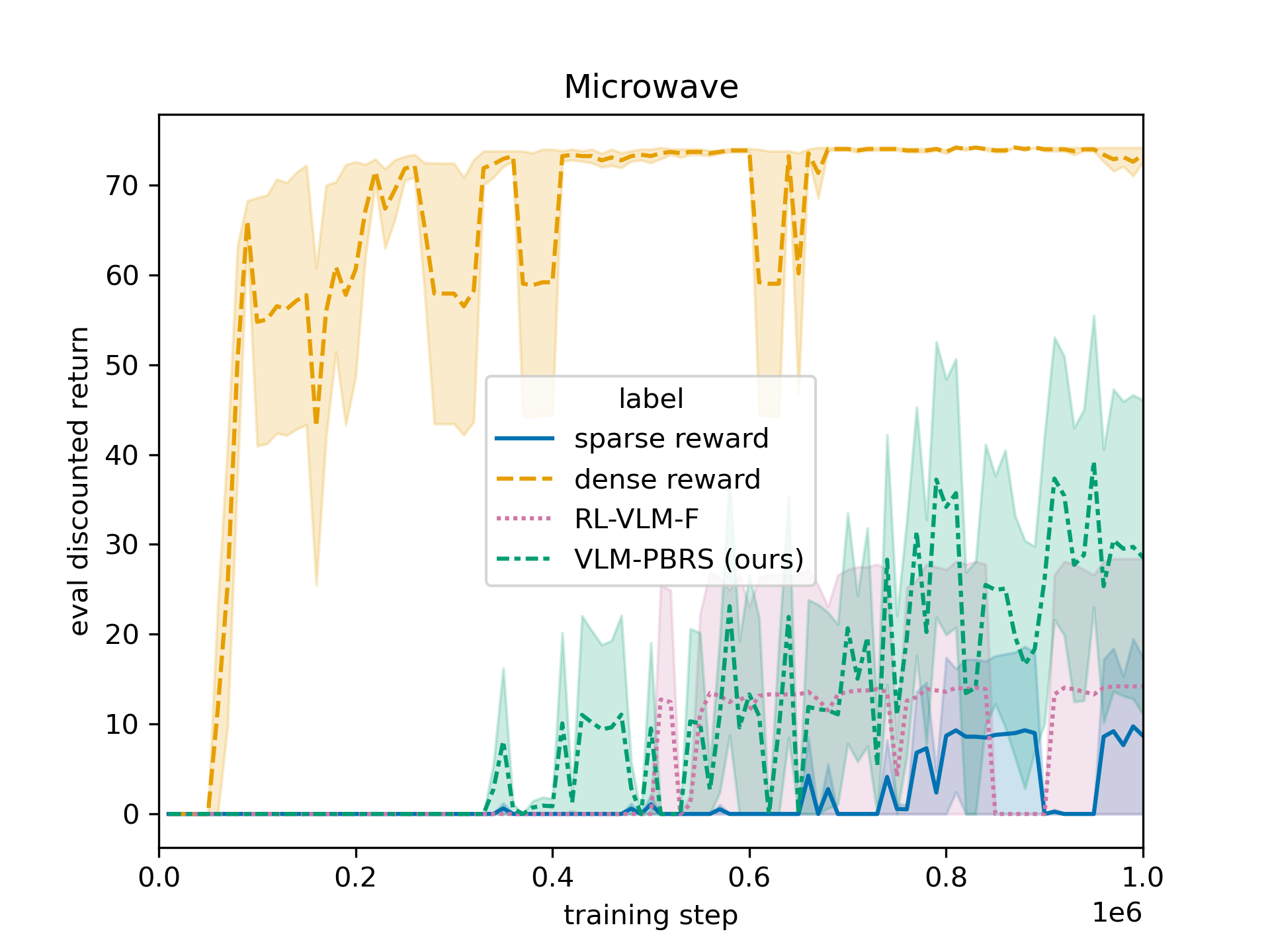}
    \caption{microwave}
\end{subfigure}
\begin{subfigure}[b]{0.35\textwidth}
    \centering
    \includegraphics[width=\textwidth]{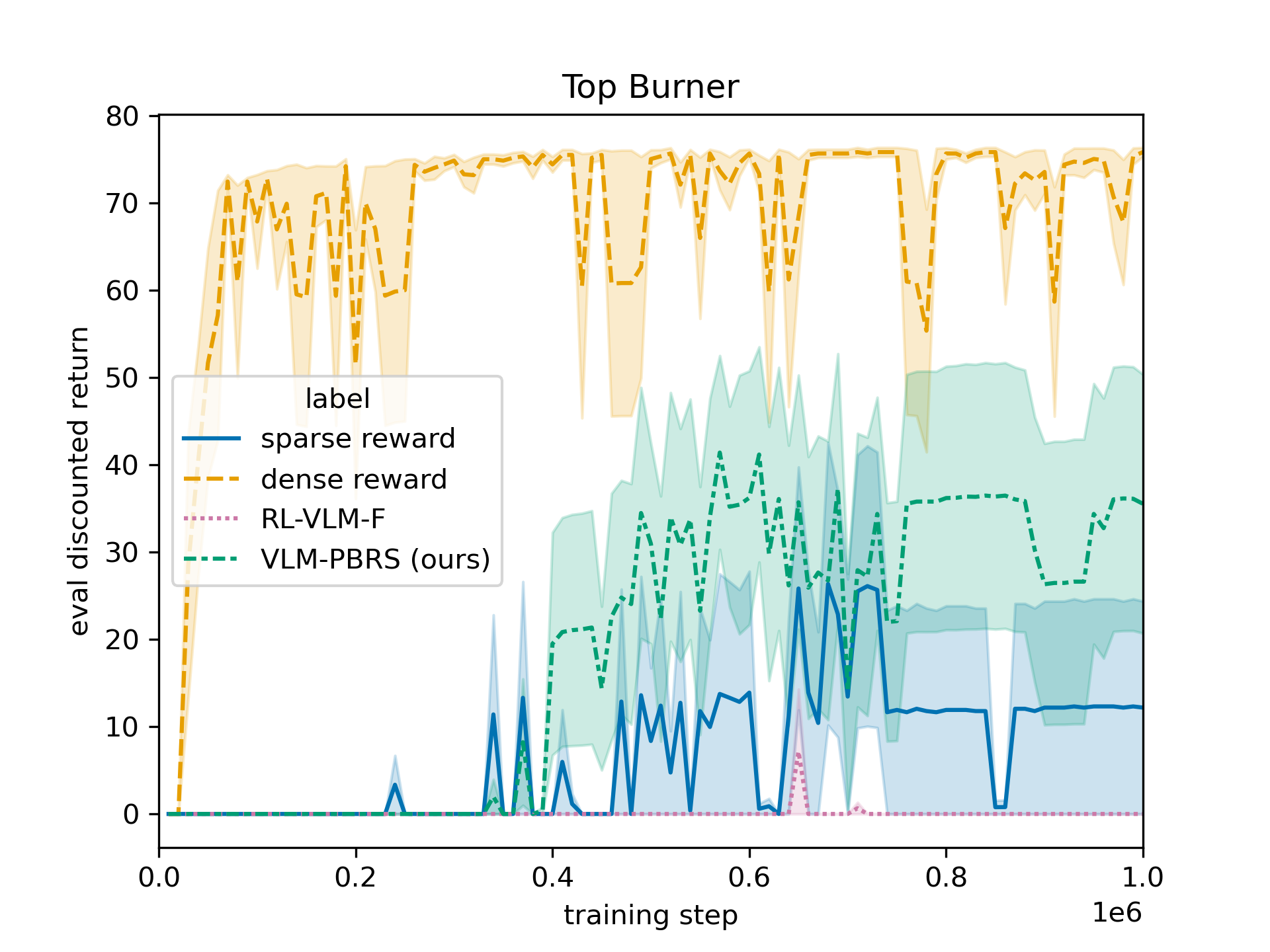}
    \caption{top-burner}
\end{subfigure} \\

\begin{subfigure}[b]{0.35\textwidth}
    \centering
    \includegraphics[width=\textwidth]{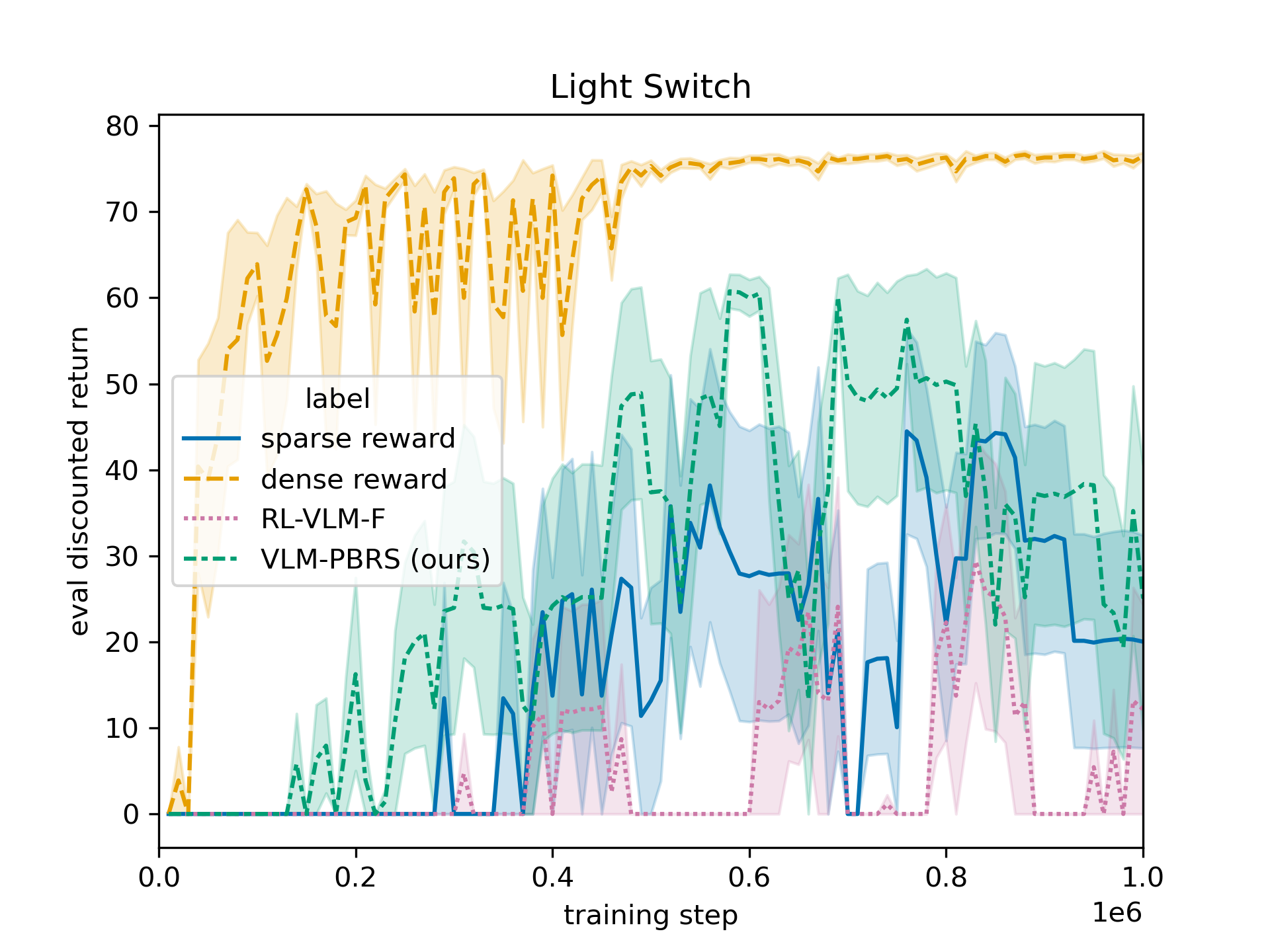}
    \caption{light-switch}
\end{subfigure} \\

\caption{Results of Meta-World (button-press, window-open) and Franka Kitchen (microwave, top-burner, light-switch). \method (in green) outperforms the dense reward baseline in button-press and improves over the sparse reward baseline in all other tasks. We report the mean and standard error of the mean of five independent, repeated training runs.}
\label{fig:results-envs}
\end{figure*}

\section{Results}

Instead of success rates, which are commonly used in Meta-World, we report the discounted episodic return on the sparse ground truth reward of evaluation episodes to measure the effect of the policy invariance of our approach on the optimality of the learned policy. Our approach not only accelerates successfully completing a task, but is able to still leverage the strength of reinforcement learning to optimize the sparse ground truth reward function when combined with incomplete or (partially) incorrect VLM feedback. For all experiments on the Meta-World and Franka Kitchen tasks, we report the mean and standard error of the mean of five independent, repeated training runs per configuration.

\subsection{Does \method Accelerate Learning?}
First, we compare the learning curves of the four methods on two example tasks of the Meta-World environment (button-press, window-open) and on three tasks of the Franka Kitchen environment (microwave, top-burner, light-switch) to evaluate whether \method can be used to automatically accelerate learning given a simple textual goal description.

The results on the learning efficiency of the methods can be found in Figure~\ref{fig:results-envs}. In each subplot the four training configurations are plotted in parallel: the ground-truth \textit{sparse} reward baseline, the human‑designed \textit{dense} reward, the VLM‑based reward learning framework \textit{RL‑VLM‑F} (added to the sparse reward), and our approach \textit{\method}. 

The ground truth sparse reward is challenging to solve with standard RL in all tasks. The average discounted returns are low and their variance is high because they depend on reaching the goal state by chance. After reaching the goal state, the optimal policy can be learned. In the repeated runs, runs that reach the goal state by chance quickly converge to the optimal policy, but in many runs the goal state will never be reached by chance, so no policy improvement can be observed, which leads to runs with discounted returns of zero over the entire training duration.

Directly using the reward function learning approach RL-VLM-F to create a denser reward improves over the sparse baseline similar to \method in the window-open, and microwave tasks, but interferes with learning well performing policies in the button-press, top-burner, and light-switch tasks.
In contrast to the sparse baseline and \methodnospace, RL-VLM-F can converge to sub-optimal policies due to errors during the VLM labeling that create incorrect incentives.

Across all tasks, \method accelerates learning over the sparse reward baseline. 
The labels generated by the small VLMs encode sufficient knowledge to accelerate learning without requiring the computational overhead associated with larger VLMs.
In the Meta-World tasks, all runs converge quickly towards the optimal policy in the early learning stages (within the first 200,000 training steps).
In the button-press task, \method is even able to consistently learn a PBRS function during the training that converges to the optimal policy faster than the given human-designed dense reward. 
In general, the human designed dense reward should be an upper bound on performance given that it builds on privileged information of the state and is given before the training starts rather than being learned during training. But, as seen in the button-press environment, as the dense reward is not potential-based and hand-crafted, it can lead to sub-optimal policies. 
The Franka Kitchen tasks are more complex due to their higher-dimensional action and state spaces and are therefore more challenging to explore and learn. 
In the Franka Kitchen tasks, \method still offers an advantage over the sparse baseline without extensive human engineering or supervision, though convergence is slower and less reliable than in the Meta-World tasks.

In summary, our approach of automatically creating a potential-based reward shaping from VLM preference labels consistently improves the sample efficiency compared to the original sparse rewards in both environments. But, the benefit of our method depends on the accuracy of the labels provided by the VLM. Notably, if the VLM provides incorrect labels, our method could also decrease the sample efficiency below the performance of the original reward, which we will explore next in Section~\ref{sec:labelquality}.

\subsection{VLM Label Quality}
\label{sec:labelquality}

\begin{table}[tb]
\centering
\resizebox{\columnwidth}{!}{
\begin{tabular}{rllllllll}
\toprule
  & button-press & window-open & drawer-open & door-open & microwave & top-burner & light-switch \\
\midrule
\textit{total} &   10,000 & 10,000 & 10,000  & 10,000 & 10,000 & 10,000 & 10,000 \\ 
\textit{pref. Image 1} &   1,906 & 639 & 567  & 1,117 & 5436 & 5356 & 3703 \\ 
\textit{pref. Image 2} &   8,094 & 9,361 & 9,335  & 8,857 & 3951 & 4081 & 5590 \\ 
\textit{no preference} &   0 & 0 & 98  & 26 & 613 & 563 & 707 \\
\textit{accuracy (excl. no pref.)} &   0.594 & 0.547 & 0.477 & 0.485 & 0.753 & 0.702 & 0.604 \\
\bottomrule
\end{tabular}
}
\caption{VLM label distributions and accuracy for all combinations and orderings of 100 sample images across the four Meta-World and three Franka Kitchen tasks.}
\label{tab:vlm-accuracy}
\end{table}

\begin{figure*}[tb]
\centering
\begin{subfigure}[b]{0.45\textwidth}
    \centering
    \begin{subfigure}[c]{0.35\textwidth}
        \includegraphics[width=\textwidth]{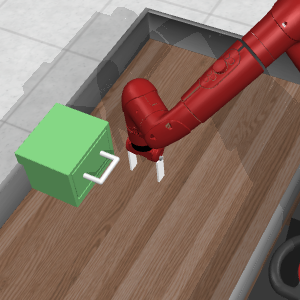}
    \end{subfigure}
    \begin{subfigure}[c]{0.6\textwidth}
        \includegraphics[width=\textwidth]{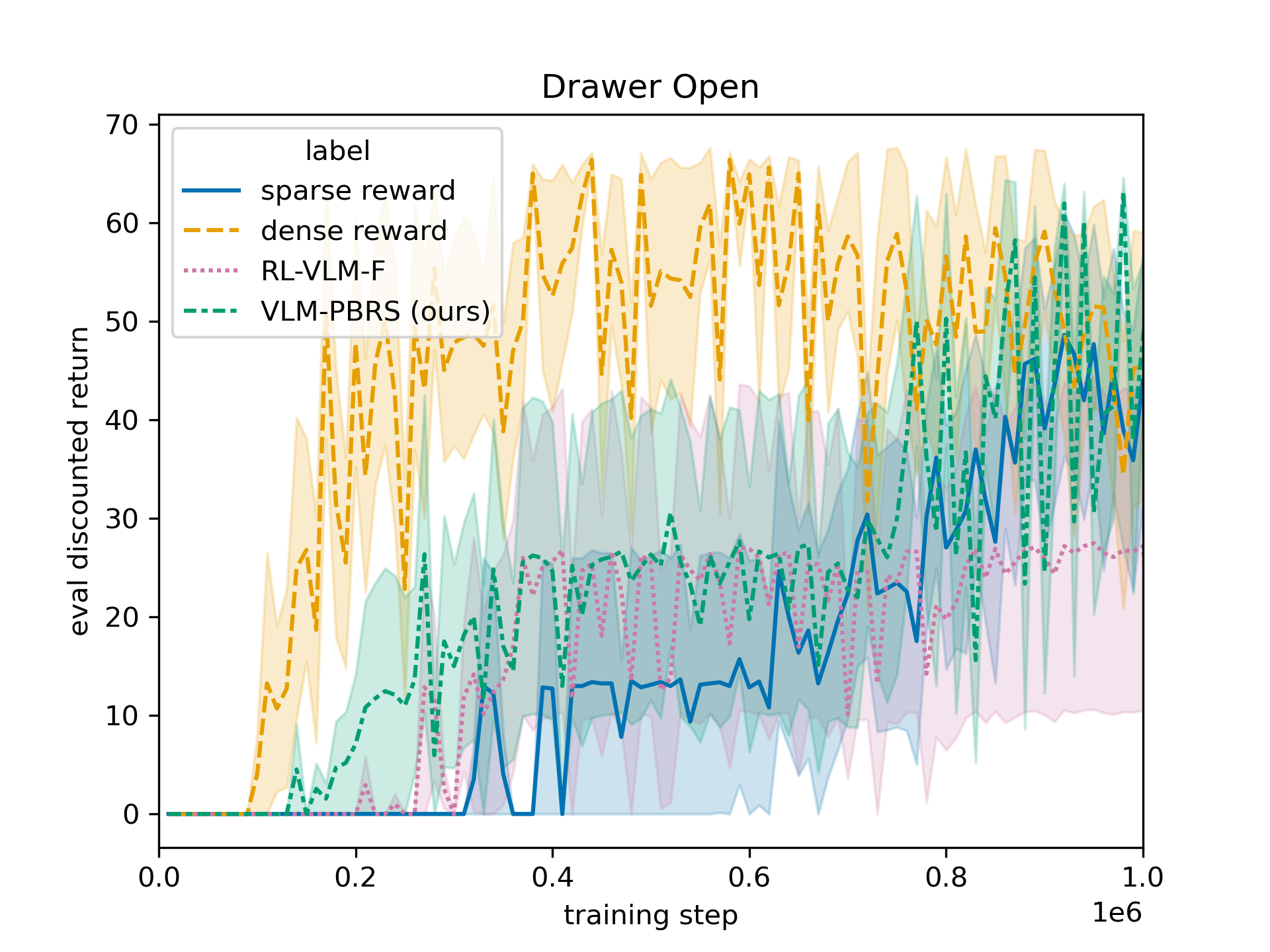}
    \end{subfigure}
     \caption{drawer-open}
 \end{subfigure}
 \begin{subfigure}[b]{0.45\textwidth}
    \centering
    \begin{subfigure}[c]{0.35\textwidth}
        \includegraphics[width=\textwidth]{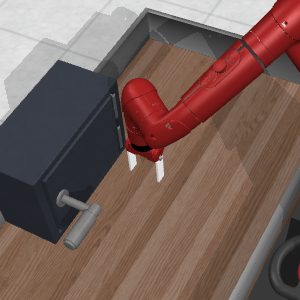}
    \end{subfigure}
    \begin{subfigure}[c]{0.6\textwidth}
        \includegraphics[width=\textwidth]{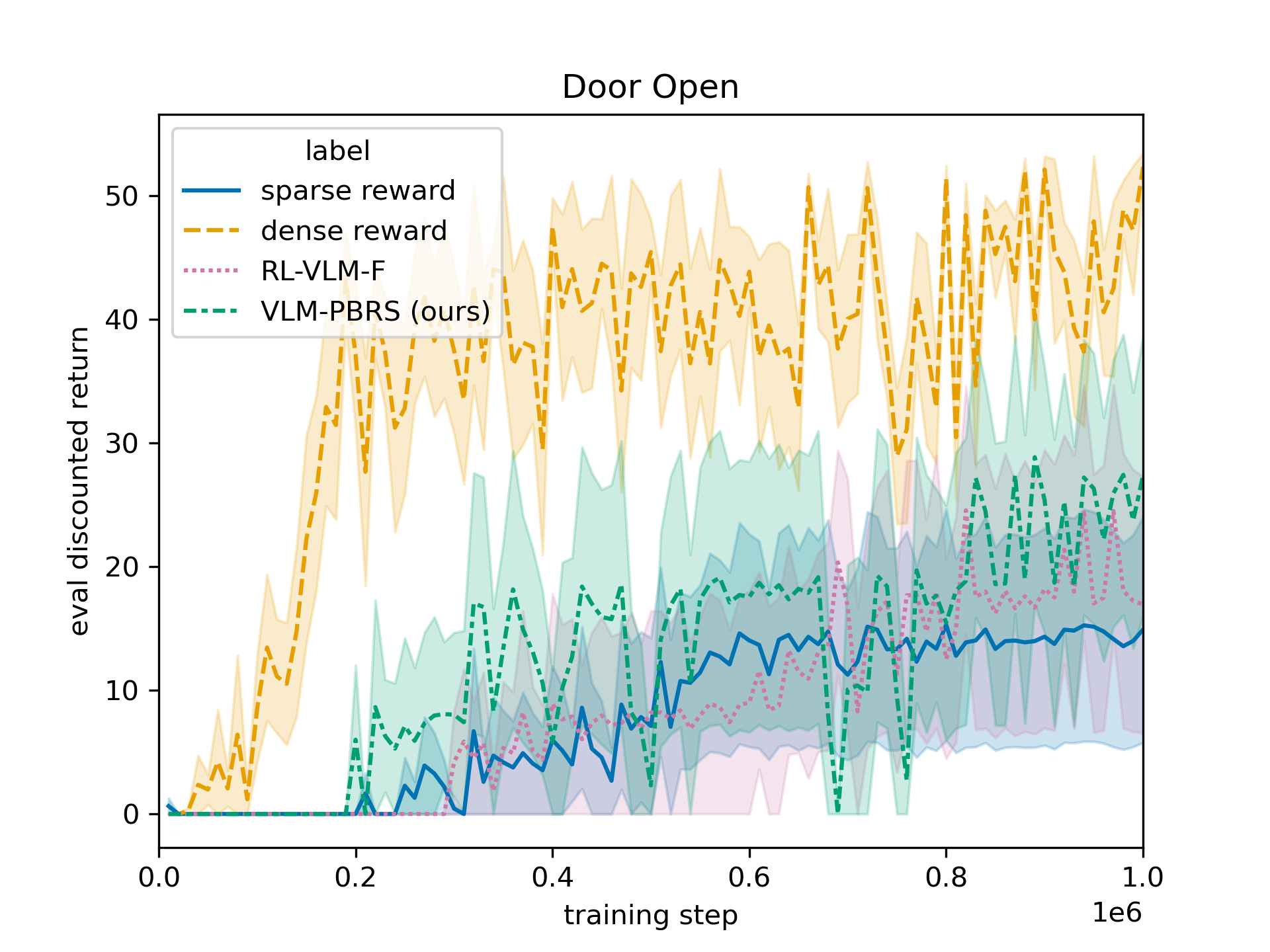} 
    \end{subfigure}
     \caption{door-open}
 \end{subfigure}
 \\

\caption{Results of tasks with at chance VLM label accuracy. \method (in green) improves slightly over the sparse baseline, but given the lack of accurate labels, fails to match the performance of the dense human-designed reward.}
\label{fig:results-low-acc-metaworld}
\end{figure*}

To evaluate the reliability of the VLM‑derived preference labels, we constructed example datasets of images that span the entire spectrum of task progress for all four tasks of Meta-World and all three Franka Kitchen tasks. 
Expert policies are executed for 50 time steps, during which the goal state is reached in each environment. This ensures that the visual cues presented to the VLM are significant representatives of realistic decision points. We sampled a total of 100 images per environment, which equates to two separate trajectories for the Meta-World tasks. 
In the Franka Kitchen tasks, the policies achieve the goal in less than 50 steps. The datasets therefore consist of three trajectories for the \textit{microwave} task and of four trajectories for the \textit{top-burner} and \textit{light-switch} tasks.
We queried the VLM for both orderings of each image pair for a total of 10000 comparisons, and therefore labels, per environment. The results can be found in Table~\ref{tab:vlm-accuracy}. We report the accuracy with regard to the ground truth labels excluding pairs labeled with \textit{no preference} by the VLM as these pairs would be ignored in reward shaping model learning.

Ground truth preference labels were computed independently for each environment. The ground truth preference consists of two parts.  First, the distance between gripper and object (or its handle). Second, the distance from the object to its goal position. 
For our calculations, we used the distance between object and goal as the primary metric. We only used the gripper-to-object distance if both states had the same object-to-goal distance (most commonly before the object was moved).

As only the comparison of each image with itself has a ground truth label of "-1", the trivial baseline chance accuracy of always preferring the first (or always preferring the second) image is $49.5\%$. 
For most tasks, the labels produced by the small VLMs surpass the baseline chance accuracy and can accelerate learning in our method as seen in Figure \ref{fig:results-envs}.
But, in the drawer-open and door-open tasks, the VLM does not perform better than trivial majority labeling. 
This may be caused by the artificial setups with the bright green drawer and the solid black door with an unusual handle shape that might be outside of the training distribution of the VLM or the issue of correctly identifying the target cabinet when presented with kitchen layout zero-shot.

Notably, Ovis2 shows a systematic order bias. Across all four Meta-World environments, at least $80\%$ of labels chosen by the VLM indicate a preference for the second image. As we test both orderings of each image pair, each image should only be preferred in half of the tests.

The no preference label "-1" is rarely used. Within each VLM fewer no preference labels are observed when the overall label accuracy is higher. In our setup, fewer no preference labels could therefore also be used as an indicator for higher label quality, though this may be specific to the used VLMs, environments, and prompt setup.

In Figure~\ref{fig:results-low-acc-metaworld}, we explore the effect of \method given low quality VLM labels, where the VLM labeling accuracy is close to random chance guessing. In both the drawer-open and door-open tasks, \method shows a slight improvement over the sparse reward baseline possibly originating from a small number of well labeled examples or more general exploration incentives. Notably, in contrast to the non-potential-based reward shaping baseline RL-VLM-F, even given the low quality labels the discounted returns do not drop below the sparse baseline.

\subsection{Ablation of VLM Preference Label Accuracy}

\begin{figure}[tb]
\centering
    \includegraphics[width=0.5\columnwidth]{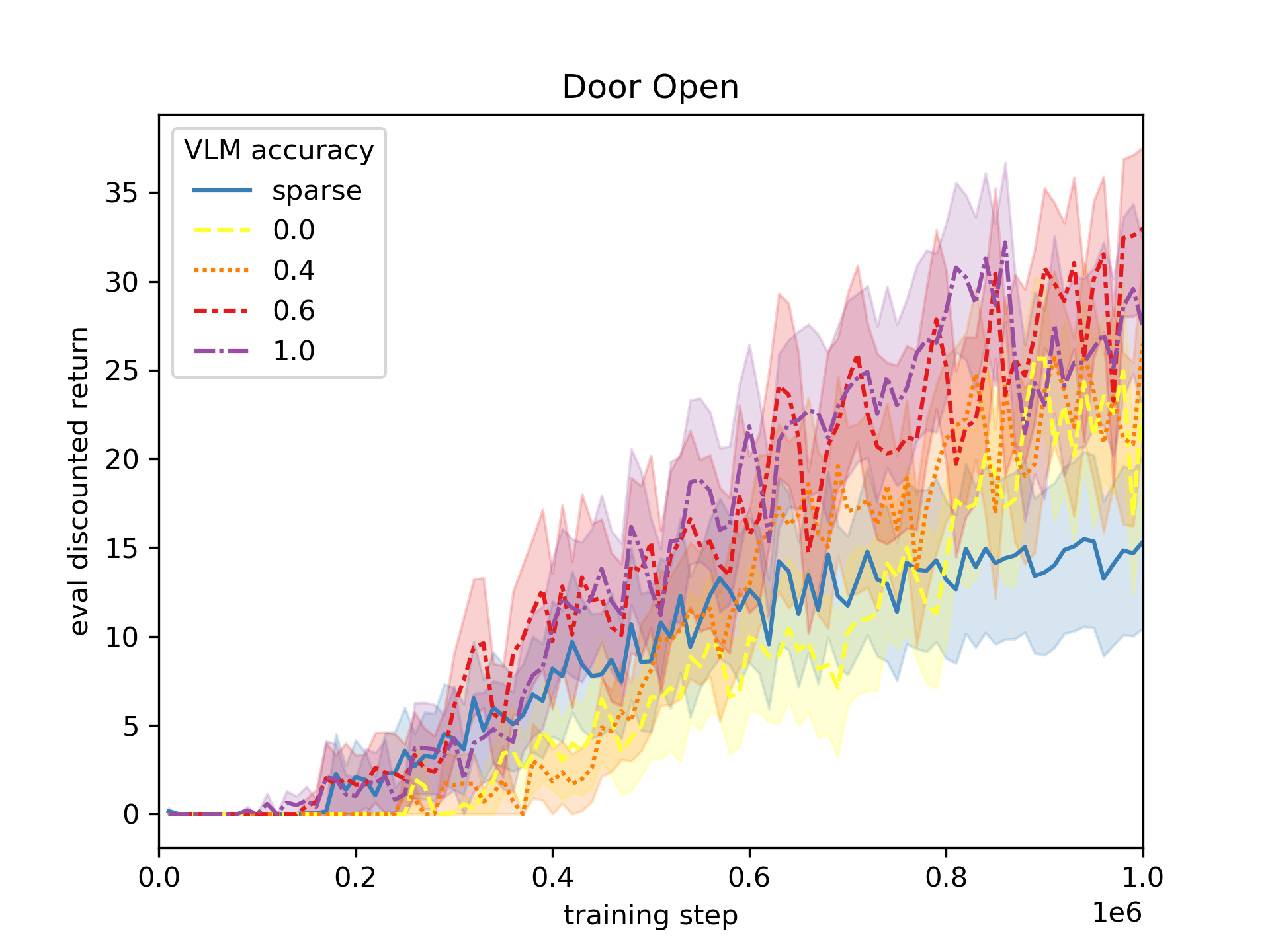}
\caption{Ablation results for fixed VLM label accuracies of an oracle labeler in \method in the door-open task of Meta-World. We report the mean discounted evaluation returns and standard error of the mean of 20 repeated training runs.}
\label{fig:ablation-accuracy-function}
\end{figure}


The VLM preference labeling task is inherently challenging, as preference labels can be ambiguous when multiple task-relevant dimensions compete. 
For instance, consider two states: one in which the agent has lost contact with the handle but the object is closer to the goal, and one in which the agent maintains a firm grip but has made less task progress. Depending on whether object control or goal proximity are valued higher, opposite labels would be assigned. Since we use two-dimensional images from a fixed position to represent the environment, it may even be impossible for the VLM to match ground truth labels due to the loss of information.


To measure the effect of VLM label quality on the convergence speed of our method, we replace the VLM with an oracle with a fixed label accuracy. 
We use the ground truth distance of gripper to target object and distance of the target object to its goal position to compute the oracle label. The preference labels are first based on the improvement of the state of the target object. If the target object is in the same state in both images, the distance of the gripper to the object handle defines the preference label. Given a target accuracy level $a$, labels are randomly flipped during the oracle labeling with a chance of $(1-a)$.

Figure~\ref{fig:ablation-accuracy-function} shows average ground truth returns for repeated runs with different target accuracy values. 
Overall, higher target accuracy values lead to quicker convergence to well performing policies. 
Accuracy levels below $0.5$ imply that the labels are likely to indicate preference for the incorrect image, which leads to a reward shaping that instead encourages to avoid achieving the goal. As a result, these runs tend to perform worse than the sparse baseline in the initial stages of learning. But given the use of PBRS, our method still learns to optimize the sparse ground truth reward even when all labels are incorrect (for $a=0.0$) relying on the inherent exploration of the underlying RL algorithm. Notably, there appears to be a positive effect on learning even with incorrect VLM feedback, where all \method runs outperform the sparse baseline. This may be caused by more efficient exploration even if the exploration does not align with the goal.
Sample efficiency improvements can be observed for better-than-chance accuracies starting with $a=0.6$. 

In general, \method will not reach the sample efficiency of a predefined dense reward function, as the shaping function has to be learned during training, but it does not require manual human effort beyond the simple high-level task description.

\subsection{Ablation of Number of VLM Preference Labels}
\begin{figure*}[tb]
\centering
    \begin{subfigure}[c]{0.4\textwidth}
        \centering
        \includegraphics[width=\textwidth]{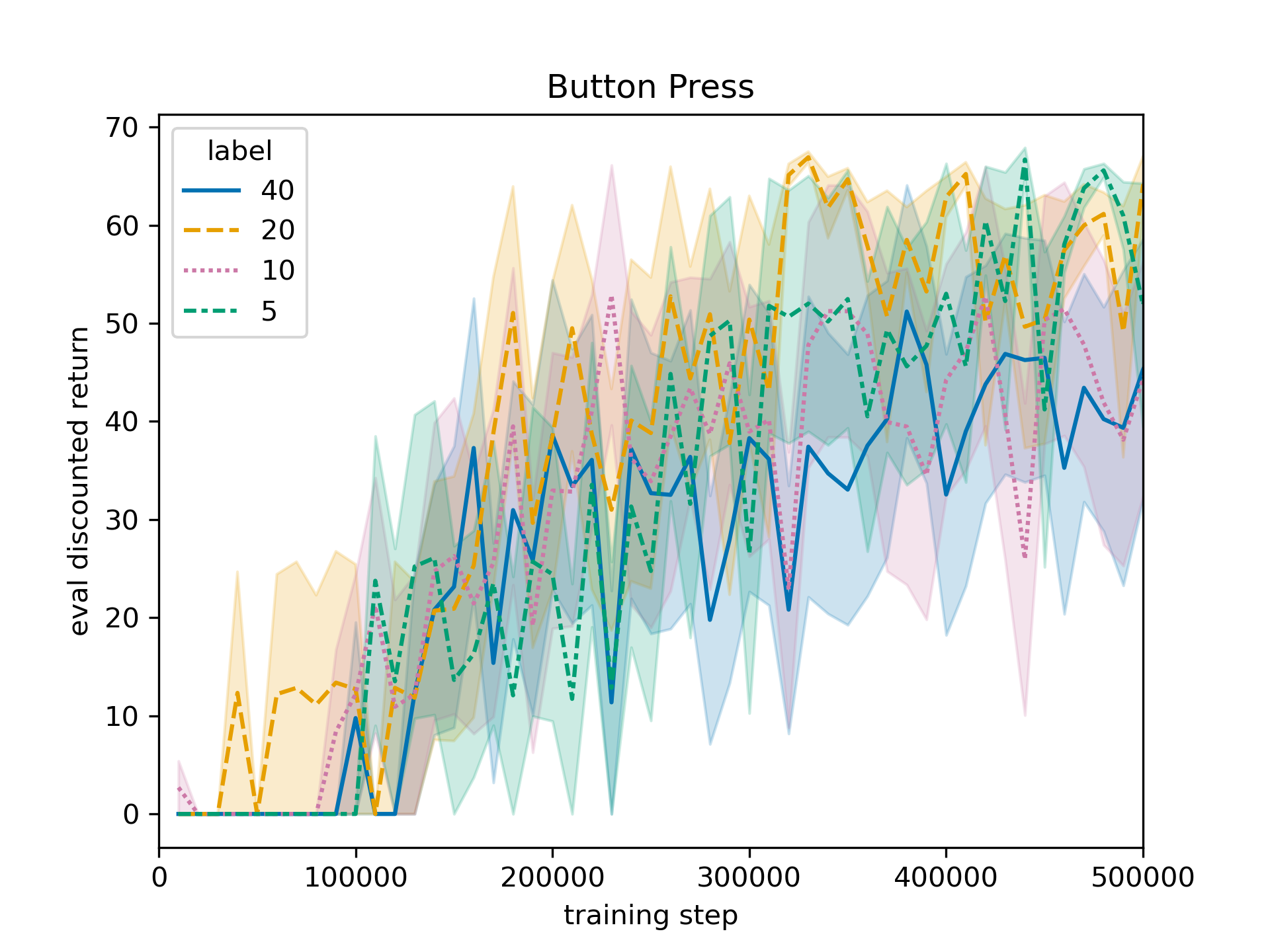}
        \caption{button-press}
    \end{subfigure}
    \begin{subfigure}[c]{0.4\textwidth}
        \centering
        \includegraphics[width=\textwidth]{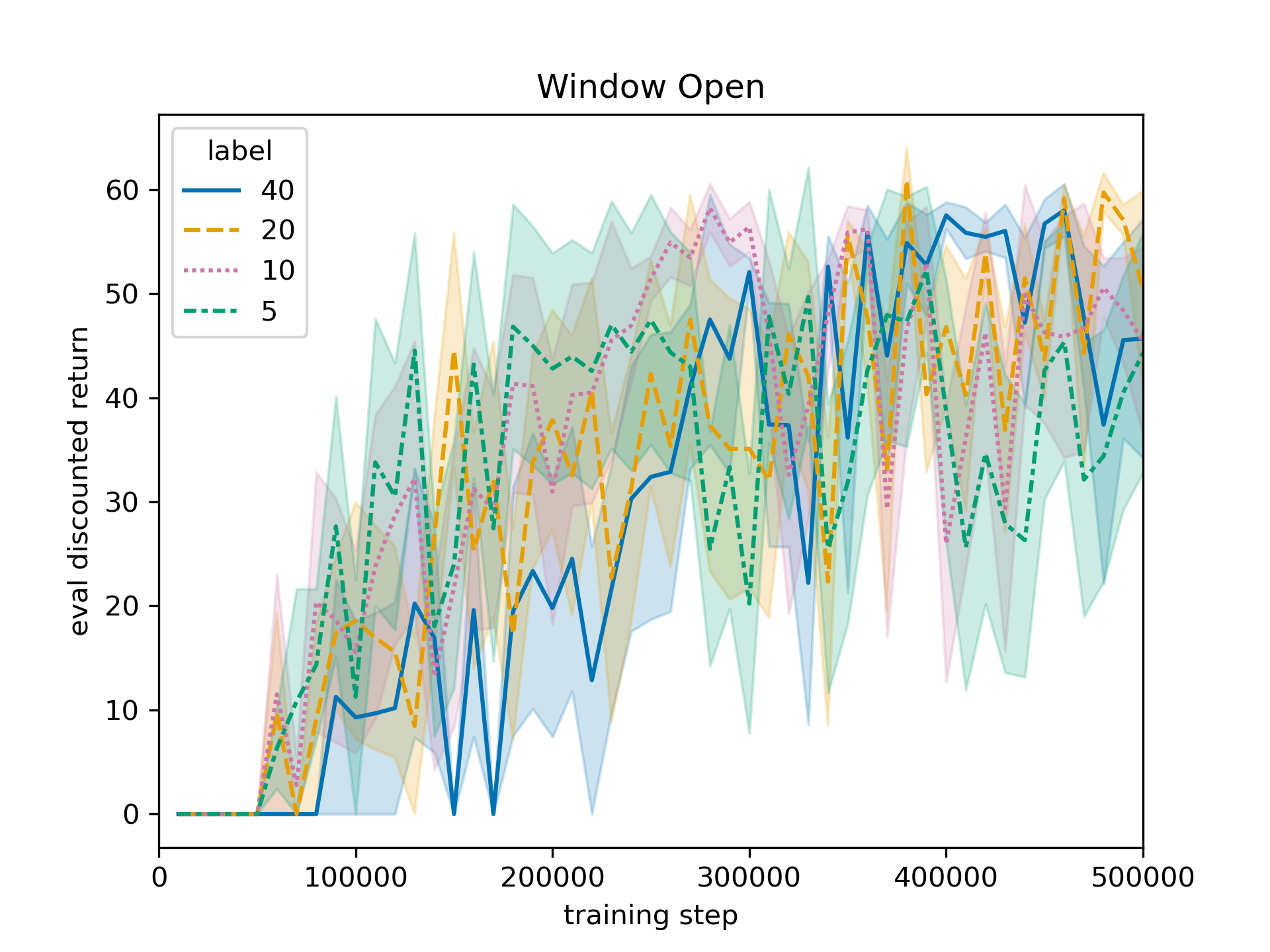}
        \caption{window-open}
    \end{subfigure}
    
    \begin{subfigure}[c]{0.4\textwidth}
        \centering
        \includegraphics[width=\textwidth]{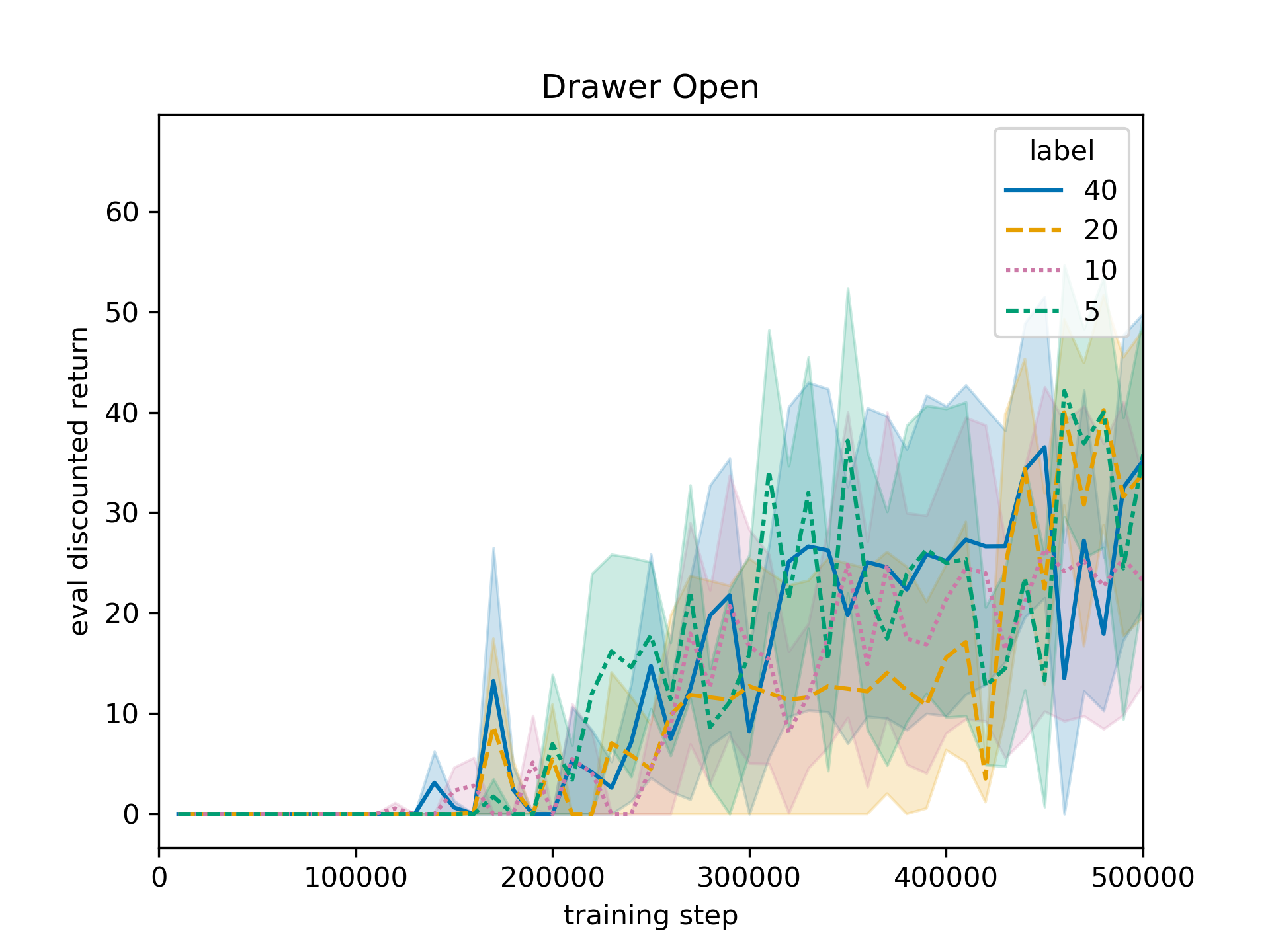}
        \caption{drawer-open}
    \end{subfigure}
    \begin{subfigure}[c]{0.4\textwidth}
        \centering
        \includegraphics[width=\textwidth]{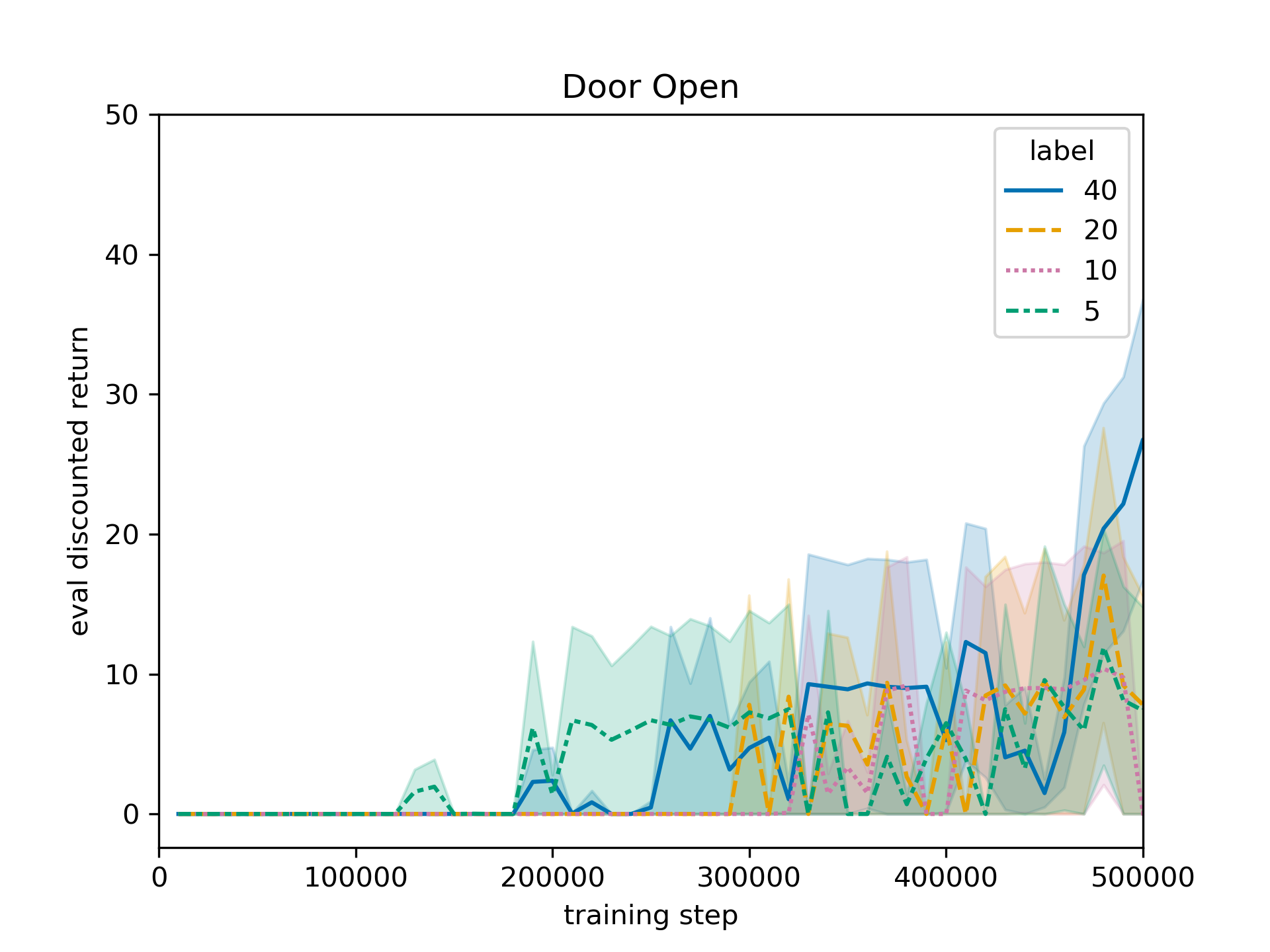}
        \caption{door-open}
    \end{subfigure}
\caption{Ablation results for number of queries per VLM labeling batch for \method in the Meta-World environment.}
\label{fig:ablation-label-number}
\end{figure*}

Figure~\ref{fig:ablation-label-number} shows the impact of the number of VLM preference labels queried in each labeling batch on learning performance for the \textit{button-press}, \textit{window-open}, \textit{drawer-open}, and \textit{door-open} tasks of Meta-World. We evaluated four different numbers of queries per VLM labeling batch: 5, 10, 20, and 40 labels per batch.

The results show that using only five labels per batch consistently yields the poorest performance across all environments. But, for larger numbers of labels per batch, little differences in sample efficiency improvements can be observed. For these choices, the mean success rates and discounted returns converge to the same asymptotic value within a similar number of steps.

Given the negligible performance gap and the lower computational overhead, we adopt 20 VLM labels per batch as our default setting in all experiments instead of the 40 VLM labels per batch used in RL-VLM-F. This choice balances efficiency with robust learning, ensuring that learning the shaping function remains effective without incurring unnecessary labeling cost.

\subsection{Ablation of Loss Function}
\label{sec:ablation-loss}

\begin{figure*}[tb]
\centering
    \begin{subfigure}[c]{0.4\textwidth}
        \centering
        \includegraphics[width=\textwidth]{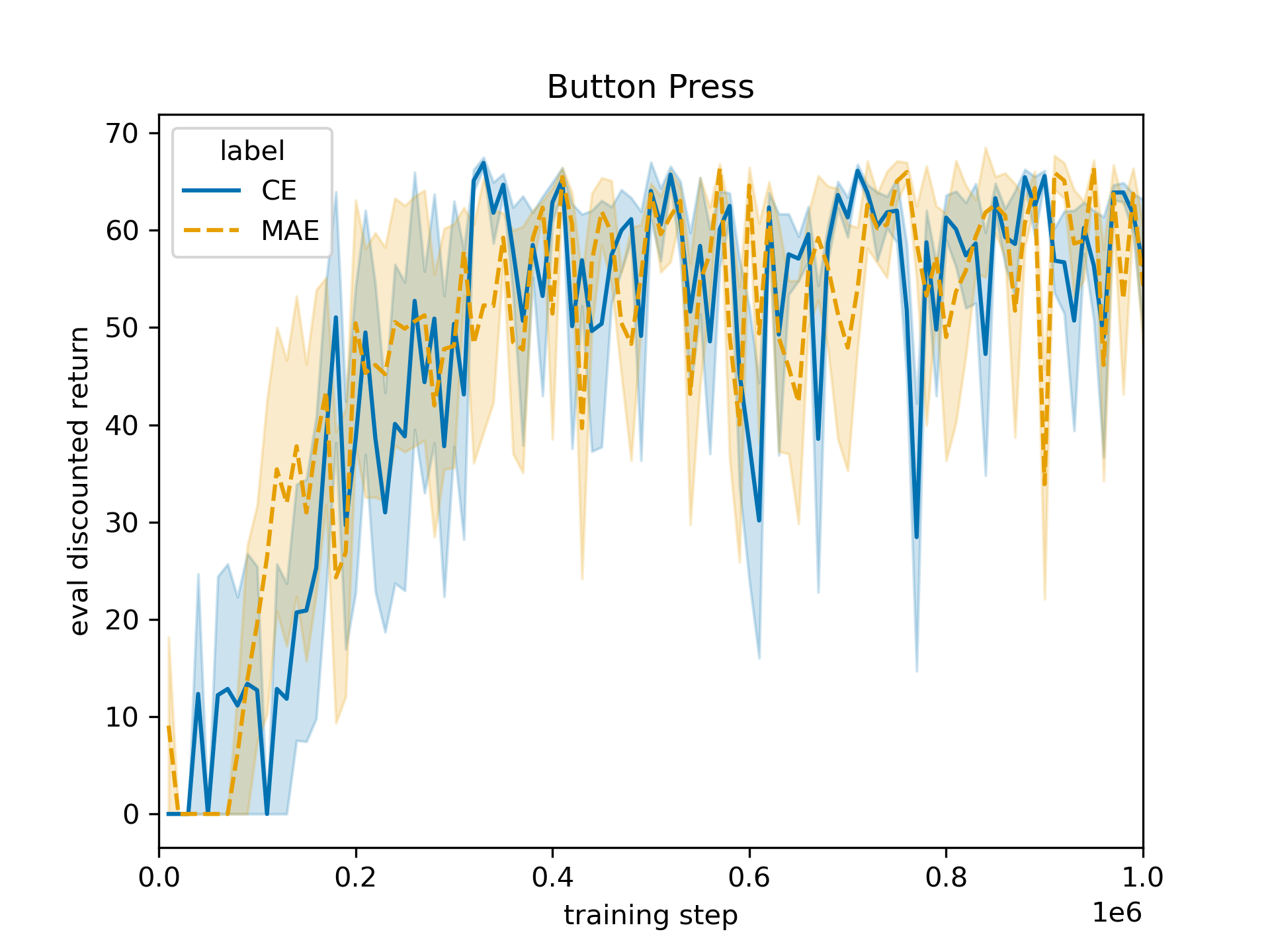}
        \caption{button-press}
    \end{subfigure}
    \begin{subfigure}[c]{0.4\textwidth}
        \centering
        \includegraphics[width=\textwidth]{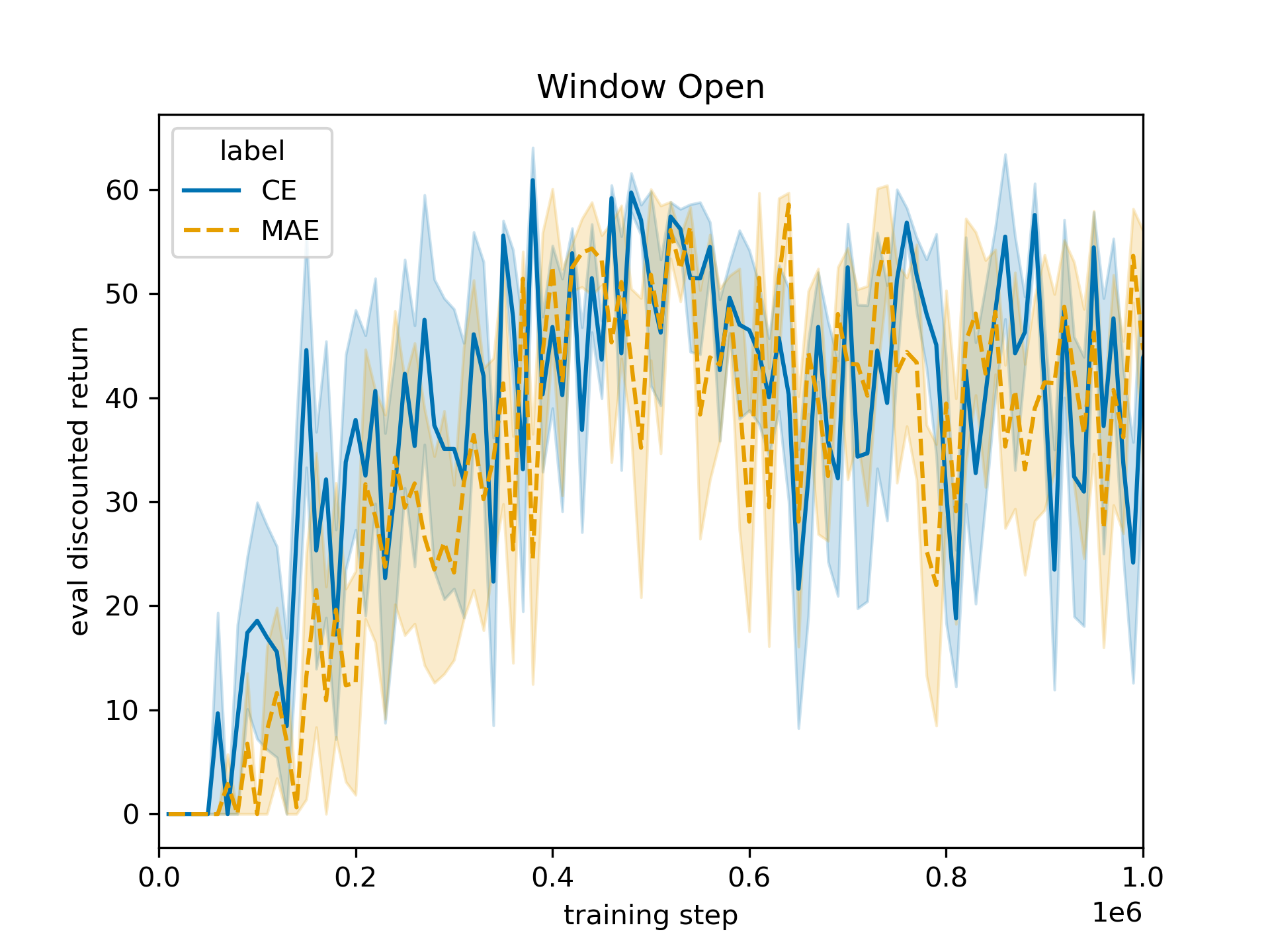}
        \caption{window-open}
    \end{subfigure}
    \begin{subfigure}[c]{0.4\textwidth}
        \centering
        \includegraphics[width=\textwidth]{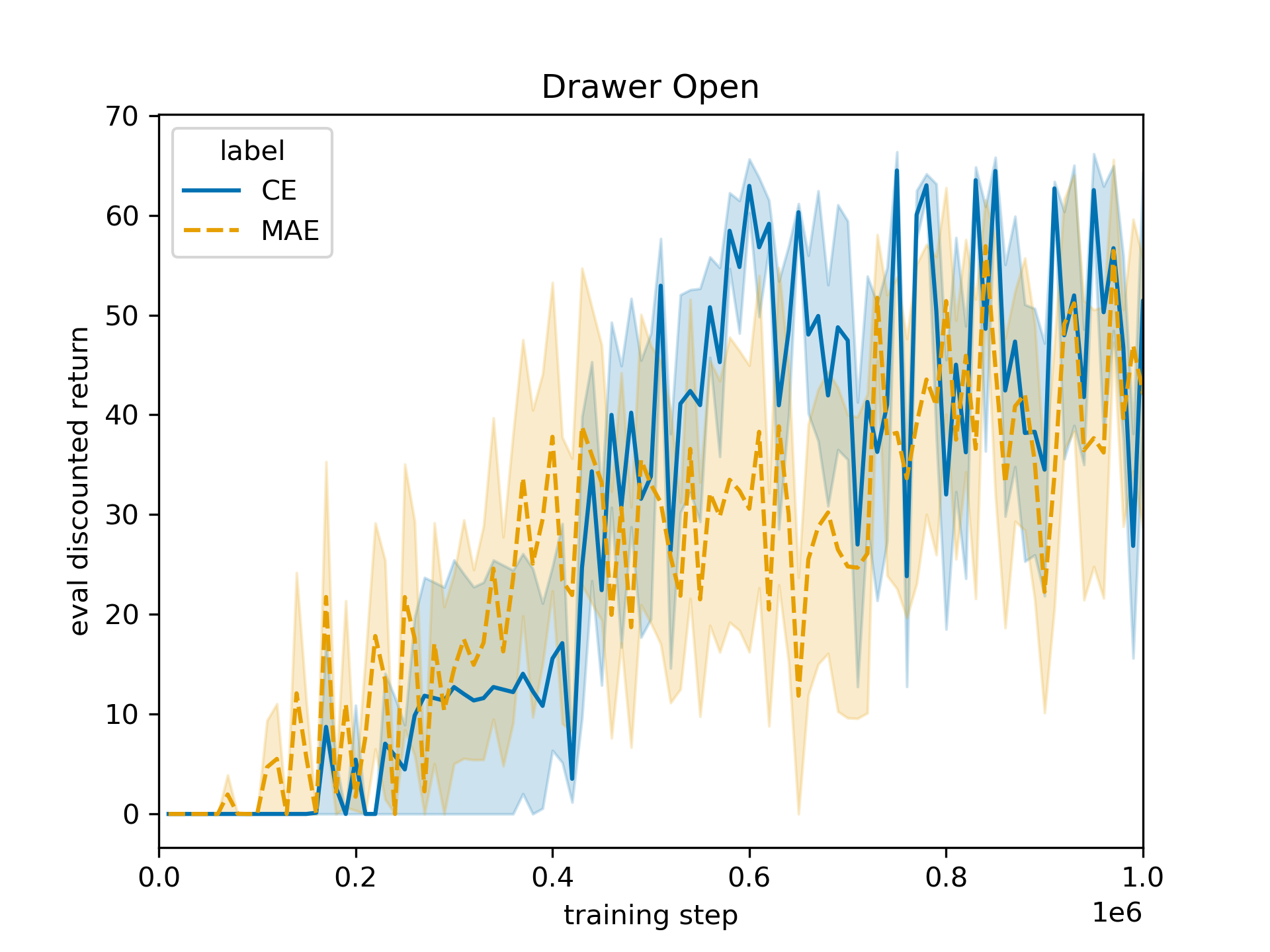}
        \caption{drawer-open}
    \end{subfigure}
    \begin{subfigure}[c]{0.4\textwidth}
        \centering
        \includegraphics[width=\textwidth]{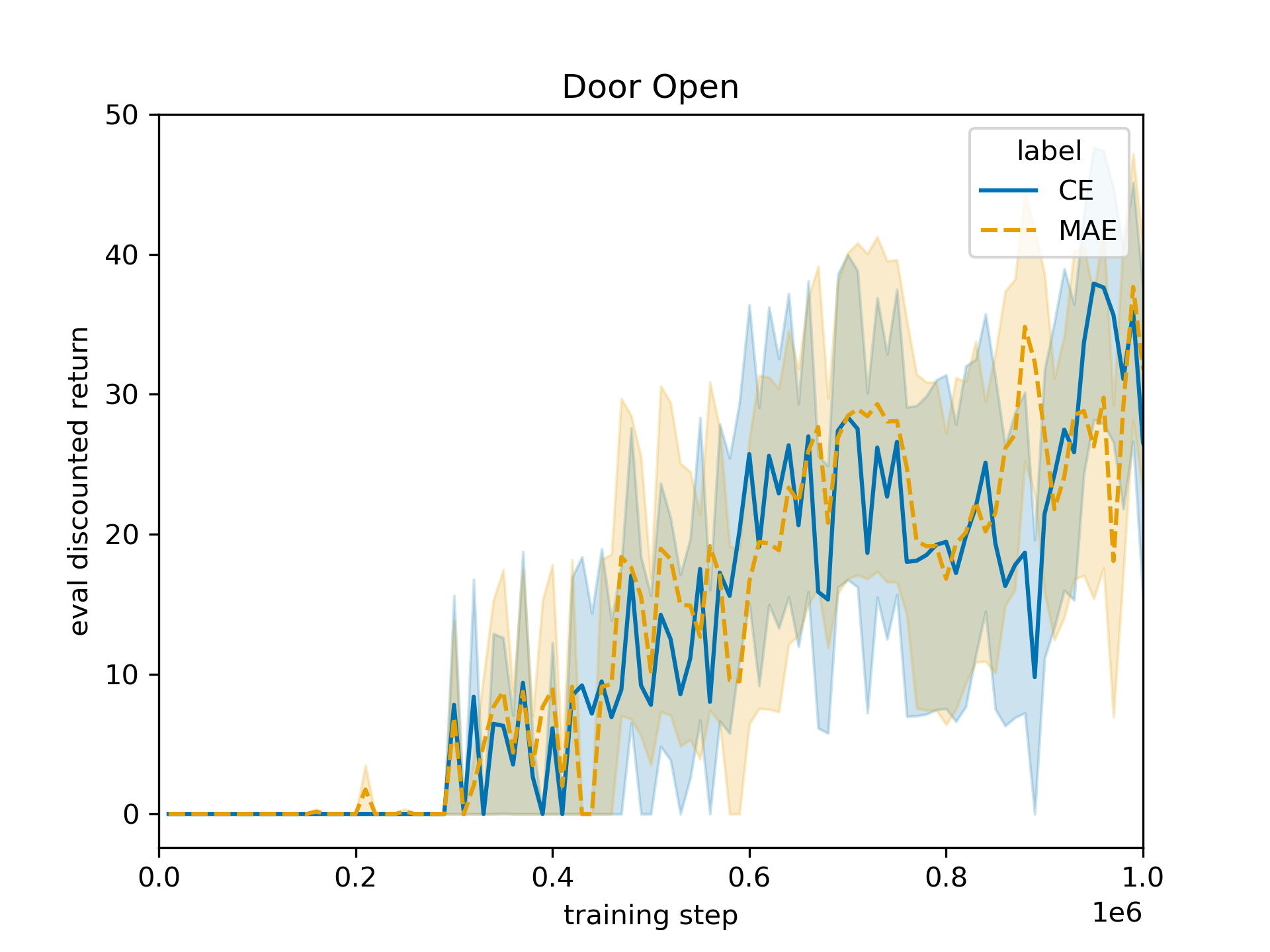}
        \caption{door-open}
    \end{subfigure}
\caption{Ablation results for choice of loss function for \method in the Meta-World environment.}
\label{fig:ablation-loss-function}
\end{figure*}

Figure~\ref{fig:ablation-loss-function} shows the results of the ablation of the loss function in \method for the \textit{button-press}, \textit{window-open}, \textit{drawer-open}, and \textit{door-open} tasks of Meta-World. Cross entropy (CE) is the default loss function used in many previous works on learning reward models from preference labels~\citep{christiano2017humanpref, lee2021pebble, wang2024rlvlmf, lin-2024-navigating-noisy, klissarov2024motif}. In \citet{ghosh2017robustloss} it was shown that mean absolute error (MAE) is more robust when learning from noisy labels, due to which it was used in \citet{luu2025ratingbased} in the context of learning reward models from VLM preference labels.

In our ablation results, both loss functions perform similar across the four tested environments, and therefore we adopted the CE loss for all other experiments.
This might be caused by the bias of the VLM to default to a preference for the second image rather than generating more arbitrarily noisy labels. As we sample the image examples uniformly at random, the label for a pair is either correct or preferring either image (as the second input image) is equally likely. 
With the CE loss, multiple repetitions of similar image pairs with flipped labels will cause the learned potential values of these pairs to move towards zero, which can cause the output distribution to move towards zero. But, this is not detrimental to our method as we focus on sparse reward settings where only the order of shaping rewards matters rather than their absolute value. In fact, the shaping rewards should be comparatively smaller than the environment sparse rewards to accelerate convergence towards optimizing the sparse, ground truth environment reward once the goal has been reached.


\section{Conclusion and Future Work}
In this work, we introduced \methodnospace, a novel approach towards automating the construction of potential-based reward shaping functions by using vision language models to accelerate reinforcement learning in sparse reward settings. 
By querying a VLM with a short textual description of the goal for preference labels over pairs of images, \method constructs a dense shaping reward guaranteed to preserve the optimal policy of the underlying sparse‑reward task. The policy invariance makes it inherently resilient against reward hacking and ensures that only the sample efficiency is influenced by the VLM label quality. This allows \method to reduce the cost of querying large VLMs and 
operate effectively with smaller VLMs (up to 16 billion parameters) and sub‑optimal labeling accuracy. The shaping signal improves exploration without altering the set of optimal policies of the original MDP, offering a practical pathway for resource‑constrained deployment.

Empirical evaluation in the Meta-World and Franka Kitchen environments demonstrated that \method consistently improves on the sparse reward baseline and outperforms the naive application of the reward learning framework RL-VLM-F~\citep{wang2024rlvlmf} as reward shaping. The sample efficiency gains correlate with VLM labeling accuracy: in the button-press environment with a sub-optimal accuracy (of 59.4\%), \method even surpasses the human‑designed dense reward baseline, indicating that automated potential-based reward shaping is a viable and scalable alternative to hand-crafted reward engineering. Both the benefit and generality of \method are expected to grow alongside the continuing progress of multi-modal foundation models.

Following this work, multiple directions for future work remain. 
The preference labeling pipeline could be further improved to enable the application for more complex tasks and ambiguous goals, and to include additional input modalities, for instance by replacing the images with (short) videos, or other richer state representations.
The example selection could also be made more principled: while the current approach biases only towards frequently visited states, a more sophisticated strategy could enable more efficient VLM labeling. 
Finally, since the potential function is updated dynamically, the shaped rewards change frequently during training, which may hinder convergence. Designing agents that inherently account for non-stationary shaping rewards thus represents a natural avenue for further improvement.


\acks{This work was supported by the Lower Saxony Ministry of Science and Culture (MWK), in the zukunft.niedersachsen program of the Volkswagen Foundation (HybrInt). The authors gratefully acknowledge the computing time granted by the KISSKI project. The calculations for this research were conducted with computing resources under the project kisski\_rl\_vlm\_f.}


\newpage


\appendix
\section{}

\subsection{Training Details}

\begin{table}[bt]
\begin{center}
\begin{tabular}{ll}
\toprule
\textbf{Hyperparameter} & \textbf{Value} \\
\midrule
Initial temperature & $0.1$ \\ 
Hidden units per layer & 256 \\ 
Number of hidden layers & 3 \\
Actor learning rate  & $0.0003$ (\textit{Meta-World}), $0.001$ (\textit{Franka Kitchen}) \\ 
Critic learning rate  & $0.0003$ \\ 
Batch Size  & $512$ \\ 
Optimizer  & Adam~\citep{kingma2014adam}\\ 
Critic target update freq & $2$ \\ 
Critic EMA $\tau$ & $0.005$ (\textit{Meta-World}), $0.001$ (\textit{Franka Kitchen}) \\ 
$(\beta_1,\beta_2)$  & $(0.9, 0.999)$ \\ 
Discount factor $\gamma$ & $0.99$\\ 
Replay buffer size & 200,000 (\textit{Rl-VLM-F, VLM-PBRS}), \\
 & 1,000,000 (\textit{sparse, dense}) \\
Initial exploration steps & 10,000 \\
\bottomrule
\end{tabular}
\end{center}
\caption{Hyperparameters of SAC used in \method and the baselines.}
\label{table:sac-hyperparameters}
\end{table}

\begin{table}[bt]
\begin{center}
\begin{tabular}{ll}
\toprule
\textbf{Hyperparameter} & \textbf{Value} \\
\midrule
Queries per VLM labeling batch $M$ & 20 \\
Frequency of VLM labeling $K$ & every 4000 steps \\
Maximum VLM query budget $N$ & 20,000 \\
Potential scaling factor $\lambda$ & $0.9$ \\
Preference Model Architecture & CNN (\textit{Meta-World}),\\
 & ResNet-18 (\textit{Franka Kitchen})\\
Output activation & sigmoid \\
Learning rate & 0.0003 \\
Image size & $300 \times 300$ (\textit{Meta-World}),\\
 & $360 \times 360$ (\textit{Franka Kitchen})\\
CNN kernel sizes & [5, 3, 3 ,3] \\
CNN channels & [16, 32, 64, 128] \\ 
CNN strides & [3, 2, 2, 2] \\
\bottomrule
\end{tabular}
\end{center}
\caption{Hyperparameters of \method and RL-VLM-F, and their preference models.}
\label{table:pbrs-hyperparameters}
\end{table}

\begin{table}[bt]
\begin{center}
\begin{tabular}{ll}
\toprule
\textbf{Hyperparameter} & \textbf{Value} \\
\midrule
model name & \texttt{Qwen3-VL-8B-Instruct} \\
greedy & false \\
top\_p  & 0.8 \\
top\_k  & 20 \\
temperature & 0.7 \\
repetition penalty & 1.0 \\
presence penalty & 1.5 \\
output sequence length & 16384 \\
\bottomrule
\end{tabular}
\end{center}
\caption{Generation hyperparameters of Qwen3-VL following the official recommendation for vision-language generation.}
\label{table:vlm-hyperparameters}
\end{table}

Tables~\ref{table:sac-hyperparameters}, \ref{table:pbrs-hyperparameters}, and~\ref{table:vlm-hyperparameters} list the default values of all hyperparameters used for \method unless explicitly changed in the ablation studies. The baselines used the same hyperparameter choices, where applicable. The hyperparameter values are based on the ones in RL-VLM-F~\citep{wang2024rlvlmf}, which builds on PEBBLE~\citep{lee2021pebble}. The learning rate and critic EMA $\tau$ were optimized in Franka Kitchen on the dense reward for more stable convergence. With the training budgets of 1,000,000 environment steps in our experiments, the maximum VLM query budget $N$ is never reached and therefore effectively unlimited.

\subsection{VLM Prompt Templates}

\begin{table}[bt]
  \centering
  \begin{tabular}{rl}
    \toprule
    Task Name & Goal Description \\ \midrule
    \textit{button-press} & Press a button. \\
    \textit{window-open} & Push and open a window. \\
    \textit{drawer-open} & Open a drawer. \\
    \textit{door-open} & Open a door with a revolving joint. \\
    \midrule
    \textit{microwave} & Open the microwave door. \\
    \textit{light-switch} & Flip the switch on the panel to turn on the stove light. \\ 
    \textit{top-burner} & Turn the oven knob that activates the top left burner. \\ 
    \bottomrule
    \end{tabular}
  \caption{Goal description used in \method and the RL-VLM-F baseline.}
  \label{tab:task_goal_description}
\end{table}

Table~\ref{tab:task_goal_description} lists the original goal descriptions for each task in Meta-World~\citep{yu2020metaworld} and partially specified goal descriptions for the Franka Kitchen~\citep{gupta2019relay, fu2020d4rl} tasks. In \textit{light-switch}, the description was extended to include the unusual position of the light switch on the panel above the stove. In \textit{top-burner}, the description was specified to explicitly state, which of the two top burners to activate.

\begin{figure*}[tb]
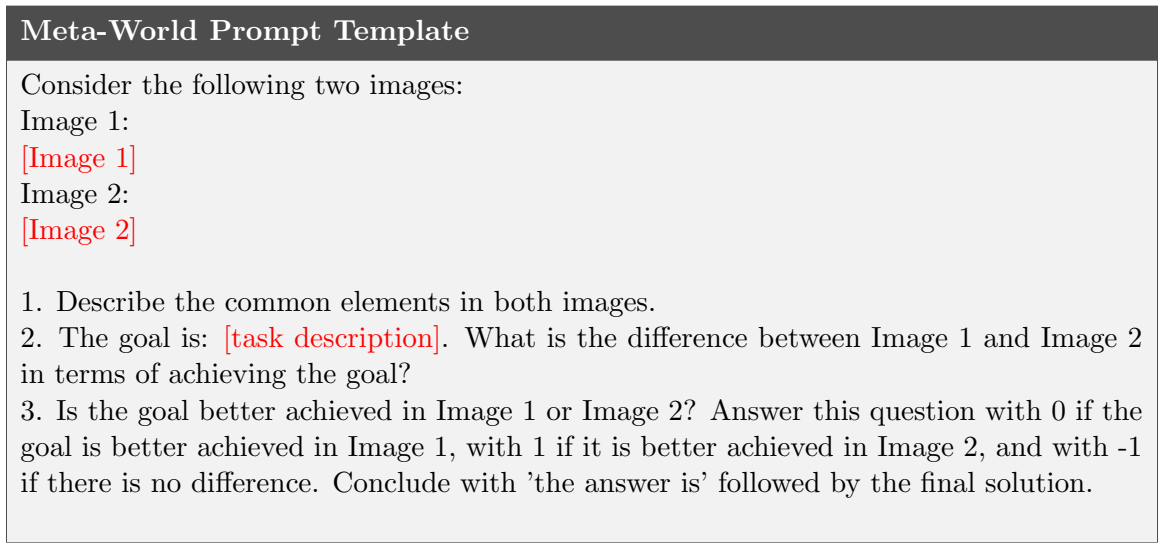

\begin{prompttextbox}[title=\textbf{Meta-World Prompt Template}]

Consider the following two images:\\
Image 1:\\
\textcolor{red}{[Image 1]}

Image 2: \\
\textcolor{red}{[Image 2]}\\

1. Describe the common elements in both images.\\
2. The goal is: \textcolor{red}{[task description]}. What is the difference between Image 1 and Image 2 in terms of achieving the goal? \\
3. Is the goal better achieved in Image 1 or Image 2? Answer this question with 0 if the goal is better achieved in Image 1, with 1 if it is better achieved in Image 2, and with -1 if there is no difference. Conclude with 'the answer is' followed by the final solution.\\

\end{prompttextbox}
\caption{The single stage prompt template for \method and the RL-VLM-F baseline used for Meta-World.}
\label{fig:prompt-ours}
\end{figure*}

These goal descriptions are then added to the templates in Figure~\ref{fig:prompt-ours} for the Meta-World environment and to the template in Figure~\ref{fig:prompt-frankakitchen} for the Franka Kitchen environment. The same prompt template is used across all tasks of one environment for both \method and RL-VLM-F. The core modification of the Franka Kitchen template is the inclusion of a system prompt with additional guidance on the labeling task.

\begin{figure*}[p]
\begin{prompttextbox}[title=\textbf{Franka Kitchen Prompt Template}]
System:\\
You are an intelligent vision-language assistant agent operating within a virtual environment. Your primary objective is to give preference labels over pairs of images given a description of a goal the robot should achieve.\\

To describe the state of target objects accurately, use spatial relationship terms like: inside, on top of, on the bottom of, on the left of, on the right of and so on. Additionally, use motion terms like open, or closed to describe the object status. To describe the object position, use the position of the object in the respective image.\\

Make sure that each description captures the essential status needed to accurately compare the images and correctly assign a preference label. The preference label should primarily be based on the state of the target object with respect to fulfilling the goal. If the target object is in the same state in both images, the preference label should be based on how close the robot gripper is to the object. If the gripper is not visible in either image, label the pair as no difference.\\

User:\\
Goal description: \textcolor{red}{[task description]}\\

Consider the following two images:\\
Image 1:\\
\textcolor{red}{[Image 1]}

Image 2: \\
\textcolor{red}{[Image 2]}\\

1. Describe the common elements in both images.\\
2. The goal is: \textcolor{red}{[task description]}. What is the difference between Image 1 and Image 2 in terms of achieving the goal? \\
3. Is the goal better achieved in Image 1 or Image 2? Answer this question with 0 if the goal is better achieved in Image 1, with 1 if it is better achieved in Image 2, and with -1 if there is no difference. Conclude with 'the answer is' followed by the final solution.\\

Think step by step before answering.\\

\end{prompttextbox}
\caption{The single stage prompt template for \method and the RL-VLM-F baseline used for Franka Kitchen.}
\label{fig:prompt-frankakitchen}
\end{figure*}

\clearpage

\vskip 0.2in
\bibliography{main}

@inproceedings{wang2024rlvlmf,
author = {Wang, Yufei and Sun, Zhanyi and Zhang, Jesse and Xian, Zhou and Biyik, Erdem and Held, David and Erickson, Zackory},
title = {RL-VLM-F: reinforcement learning from vision language foundation model feedback},
year = {2024},
publisher = {JMLR.org},
booktitle = {Proceedings of the 41st International Conference on Machine Learning},
articleno = {2112},
numpages = {18},
location = {Vienna, Austria},
series = {ICML'24}
}

@inproceedings{christiano2017humanpref,
 author = {Christiano, Paul F and Leike, Jan and Brown, Tom and Martic, Miljan and Legg, Shane and Amodei, Dario},
 booktitle = {Advances in Neural Information Processing Systems},
 editor = {I. Guyon and U. Von Luxburg and S. Bengio and H. Wallach and R. Fergus and S. Vishwanathan and R. Garnett},
 pages = {},
 publisher = {Curran Associates, Inc.},
 title = {Deep Reinforcement Learning from Human Preferences},
 url = {https://proceedings.neurips.cc/paper_files/paper/2017/file/d5e2c0adad503c91f91df240d0cd4e49-Paper.pdf},
 volume = {30},
 year = {2017}
}

@article{wirth2017prefrl,
  author  = {Christian Wirth and Riad Akrour and Gerhard Neumann and Johannes F{{\"u}}rnkranz},
  title   = {A Survey of Preference-Based Reinforcement Learning Methods},
  journal = {Journal of Machine Learning Research},
  year    = {2017},
  volume  = {18},
  number  = {136},
  pages   = {1--46},
  url     = {http://jmlr.org/papers/v18/16-634.html}
}

@InProceedings{lee2021pebble,
  title = 	 {PEBBLE: Feedback-Efficient Interactive Reinforcement Learning via Relabeling Experience and Unsupervised Pre-training},
  author =       {Lee, Kimin and Smith, Laura M and Abbeel, Pieter},
  booktitle = 	 {Proceedings of the 38th International Conference on Machine Learning},
  pages = 	 {6152--6163},
  year = 	 {2021},
  editor = 	 {Meila, Marina and Zhang, Tong},
  volume = 	 {139},
  series = 	 {Proceedings of Machine Learning Research},
  month = 	 {18--24 Jul},
  publisher =    {PMLR},
  url = 	 {https://proceedings.mlr.press/v139/lee21i.html}
}

@article{lu2024ovis,
  title={Ovis: Structural Embedding Alignment for Multimodal Large Language Model},
  author={Shiyin Lu and Yang Li and Qing-Guo Chen and Zhao Xu and Weihua Luo and Kaifu Zhang and Han-Jia Ye},
  year={2024},
  journal={arXiv:2405.20797}
}

@article{yang2025qwen3,
    title={Qwen3 Technical Report}, 
    author={An Yang and Anfeng Li and Baosong Yang and Beichen Zhang and Binyuan Hui and Bo Zheng and Bowen Yu and Chang Gao and Chengen Huang and Chenxu Lv and Chujie Zheng and Dayiheng Liu and Fan Zhou and Fei Huang and Feng Hu and Hao Ge and Haoran Wei and Huan Lin and Jialong Tang and Jian Yang and Jianhong Tu and Jianwei Zhang and Jianxin Yang and Jiaxi Yang and Jing Zhou and Jingren Zhou and Junyang Lin and Kai Dang and Keqin Bao and Kexin Yang and Le Yu and Lianghao Deng and Mei Li and Mingfeng Xue and Mingze Li and Pei Zhang and Peng Wang and Qin Zhu and Rui Men and Ruize Gao and Shixuan Liu and Shuang Luo and Tianhao Li and Tianyi Tang and Wenbiao Yin and Xingzhang Ren and Xinyu Wang and Xinyu Zhang and Xuancheng Ren and Yang Fan and Yang Su and Yichang Zhang and Yinger Zhang and Yu Wan and Yuqiong Liu and Zekun Wang and Zeyu Cui and Zhenru Zhang and Zhipeng Zhou and Zihan Qiu},
    journal = {arXiv preprint arXiv:2505.09388},
    year={2025}
}

@InProceedings{yu2020metaworld,
  title = 	 {Meta-World: A Benchmark and Evaluation for Multi-Task and Meta Reinforcement Learning},
  author =       {Yu, Tianhe and Quillen, Deirdre and He, Zhanpeng and Julian, Ryan and Hausman, Karol and Finn, Chelsea and Levine, Sergey},
  booktitle = 	 {Proceedings of the Conference on Robot Learning},
  pages = 	 {1094--1100},
  year = 	 {2020},
  editor = 	 {Kaelbling, Leslie Pack and Kragic, Danica and Sugiura, Komei},
  volume = 	 {100},
  series = 	 {Proceedings of Machine Learning Research},
  month = 	 {30 Oct--01 Nov},
  publisher =    {PMLR},
  url = 	 {https://proceedings.mlr.press/v100/yu20a.html},
}

@article{gupta2019relay,
  title={Relay policy learning: Solving long-horizon tasks via imitation and reinforcement learning},
  author={Gupta, Abhishek and Kumar, Vikash and Lynch, Corey and Levine, Sergey and Hausman, Karol},
  journal={arXiv preprint arXiv:1910.11956},
  year={2019}
}

@misc{fu2020d4rl,
    title={D4RL: Datasets for Deep Data-Driven Reinforcement Learning},
    author={Justin Fu and Aviral Kumar and Ofir Nachum and George Tucker and Sergey Levine},
    year={2020},
    eprint={2004.07219},
    archivePrefix={arXiv},
    primaryClass={cs.LG}
}

@InProceedings{haarnoja2018sac,
  title = 	 {Soft Actor-Critic: Off-Policy Maximum Entropy Deep Reinforcement Learning with a Stochastic Actor},
  author =       {Haarnoja, Tuomas and Zhou, Aurick and Abbeel, Pieter and Levine, Sergey},
  booktitle = 	 {Proceedings of the 35th International Conference on Machine Learning},
  pages = 	 {1861--1870},
  year = 	 {2018},
  editor = 	 {Dy, Jennifer and Krause, Andreas},
  volume = 	 {80},
  series = 	 {Proceedings of Machine Learning Research},
  month = 	 {10--15 Jul},
  publisher =    {PMLR},
  url = 	 {https://proceedings.mlr.press/v80/haarnoja18b.html}
}

@inproceedings{kingma2014adam,
  title={Adam: A method for stochastic optimization},
  author={Kingma, Diederik P and Ba, Jimmy},
  booktitle={International Conference on Learning Representations},
  year={2015}
}

@inproceedings{lin-2024-navigating-noisy,
    title = "Navigating Noisy Feedback: Enhancing Reinforcement Learning with Error-Prone Language Models",
    author = "Lin, Muhan  and
      Shi, Shuyang  and
      Guo, Yue  and
      Chalaki, Behdad  and
      Tadiparthi, Vaishnav  and
      Moradi Pari, Ehsan  and
      Stepputtis, Simon  and
      Campbell, Joseph  and
      Sycara, Katia P.",
    editor = "Al-Onaizan, Yaser  and
      Bansal, Mohit  and
      Chen, Yun-Nung",
    booktitle = "Findings of the Association for Computational Linguistics: EMNLP 2024",
    month = nov,
    year = "2024",
    address = "Miami, Florida, USA",
    publisher = "Association for Computational Linguistics",
    url = "https://aclanthology.org/2024.findings-emnlp.939/",
    doi = "10.18653/v1/2024.findings-emnlp.939",
    pages = "16002--16014"
}

@inproceedings{
chan2023visionlanguage,
title={Vision-Language Models as a Source of Rewards},
author={Harris Chan and Volodymyr Mnih and Feryal Behbahani and Michael Laskin and Luyu Wang and Fabio Pardo and Maxime Gazeau and Himanshu Sahni and Dan Horgan and Kate Baumli and Yannick Schroecker and Stephen Spencer and Richie Steigerwald and John Quan and Gheorghe Comanici and Sebastian Flennerhag and Alexander Neitz and Lei M Zhang and Tom Schaul and Satinder Singh and Clare Lyle and Tim Rockt{\"a}schel and Jack Parker-Holder and Kristian Holsheimer},
booktitle={Second Agent Learning in Open-Endedness Workshop},
year={2023},
url={https://openreview.net/forum?id=Xw1hVTWxxQ}
}

@inproceedings{
klissarov2025modelingcapabilities,
title={On the Modeling Capabilities of Large Language Models for Sequential Decision Making},
author={Martin Klissarov and R Devon Hjelm and Alexander T Toshev and Bogdan Mazoure},
booktitle={The Thirteenth International Conference on Learning Representations},
year={2025},
url={https://openreview.net/forum?id=vodsIF3o7N}
}

@inproceedings{
    luu2025ratingbased,
    title={Enhancing Rating-Based Reinforcement Learning to Effectively Leverage Feedback from Large Vision-Language Models},
    author={Tung Minh Luu and Younghwan Lee and Donghoon Lee and Sunho Kim and Min Jun Kim and Chang D. Yoo},
    booktitle={Forty-second International Conference on Machine Learning},
    year={2025},
    url={https://openreview.net/forum?id=k77bq8AJVy}
}

@inproceedings{ghosh2017robustloss,
author = {Ghosh, Aritra and Kumar, Himanshu and Sastry, P. S.},
title = {Robust loss functions under label noise for deep neural networks},
year = {2017},
publisher = {AAAI Press},
booktitle = {Proceedings of the Thirty-First AAAI Conference on Artificial Intelligence},
pages = {1919–1925},
numpages = {7},
location = {San Francisco, California, USA},
series = {AAAI'17}
}

@InProceedings{zhao2025consistent,
    author    = {Zhao, Yinuo and Yuan, Jiale and Xu, Zhiyuan and Hao, Xiaoshuai and Zhang, Xinyi and Wu, Kun and Che, Zhengping and Liu, Chi Harold and Tang, Jian},
    title     = {Training-free Generation of Temporally Consistent Rewards from VLMs},
    booktitle = {Proceedings of the IEEE/CVF International Conference on Computer Vision (ICCV)},
    month     = {October},
    year      = {2025},
    pages     = {8133-8143}
}

@article{bradley1952rank,
  title={Rank analysis of incomplete block designs: I. The method of paired comparisons},
  author={Bradley, Ralph Allan and Terry, Milton E},
  journal={Biometrika},
  volume={39},
  number={3/4},
  pages={324--345},
  year={1952},
}

@inproceedings{randlov1998bicycle,
author = {Randl\o{}v, Jette and Alstr\o{}m, Preben},
title = {Learning to Drive a Bicycle Using Reinforcement Learning and Shaping},
year = {1998},
isbn = {1558605568},
publisher = {Morgan Kaufmann Publishers Inc.},
address = {San Francisco, CA, USA},
booktitle = {Proceedings of the Fifteenth International Conference on Machine Learning},
pages = {463–471},
numpages = {9},
series = {ICML '98}
}

@inproceedings{hu2020utilize,
 author = {Hu, Yujing and Wang, Weixun and Jia, Hangtian and Wang, Yixiang and Chen, Yingfeng and Hao, Jianye and Wu, Feng and Fan, Changjie},
 booktitle = {Advances in Neural Information Processing Systems},
 editor = {H. Larochelle and M. Ranzato and R. Hadsell and M.F. Balcan and H. Lin},
 pages = {15931--15941},
 publisher = {Curran Associates, Inc.},
 title = {Learning to Utilize Shaping Rewards: A New Approach of Reward Shaping},
 url = {https://proceedings.neurips.cc/paper_files/paper/2020/file/b710915795b9e9c02cf10d6d2bdb688c-Paper.pdf},
 volume = {33},
 year = {2020}
}

@INPROCEEDINGS{memarian2021selfsupervisedrs,
  author={Memarian, Farzan and Goo, Wonjoon and Lioutikov, Rudolf and Niekum, Scott and Topcu, Ufuk},
  booktitle={2021 IEEE/RSJ International Conference on Intelligent Robots and Systems (IROS)}, 
  title={Self-Supervised Online Reward Shaping in Sparse-Reward Environments}, 
  year={2021},
  volume={},
  number={},
  pages={2369-2375},
  doi={10.1109/IROS51168.2021.9636020}
}

@inproceedings{devidze2022explorationrs,
 author = {Devidze, Rati and Kamalaruban, Parameswaran and Singla, Adish},
 booktitle = {Advances in Neural Information Processing Systems},
 editor = {S. Koyejo and S. Mohamed and A. Agarwal and D. Belgrave and K. Cho and A. Oh},
 pages = {5829--5842},
 publisher = {Curran Associates, Inc.},
 title = {Exploration-Guided Reward Shaping for Reinforcement Learning under Sparse Rewards},
 url = {https://proceedings.neurips.cc/paper_files/paper/2022/file/266c0f191b04cbbbe529016d0edc847e-Paper-Conference.pdf},
 volume = {35},
 year = {2022}
}

@inproceedings{trott2019keepingyourdistance,
 author = {Trott, Alexander and Zheng, Stephan and Xiong, Caiming and Socher, Richard},
 booktitle = {Advances in Neural Information Processing Systems},
 editor = {H. Wallach and H. Larochelle and A. Beygelzimer and F. d\textquotesingle Alch\'{e}-Buc and E. Fox and R. Garnett},
 pages = {},
 publisher = {Curran Associates, Inc.},
 title = {Keeping Your Distance: Solving Sparse Reward Tasks Using Self-Balancing Shaped Rewards},
 url = {https://proceedings.neurips.cc/paper_files/paper/2019/file/64c26b2a2dcf068c49894bd07e0e6389-Paper.pdf},
 volume = {32},
 year = {2019}
}

@inproceedings{
klissarov2024motif,
title={Motif: Intrinsic Motivation from Artificial Intelligence Feedback},
author={Martin Klissarov and Pierluca D'Oro and Shagun Sodhani and Roberta Raileanu and Pierre-Luc Bacon and Pascal Vincent and Amy Zhang and Mikael Henaff},
booktitle={The Twelfth International Conference on Learning Representations},
year={2024},
url={https://openreview.net/forum?id=tmBKIecDE9}
}

@article{chu2023accelerating,
  title={Accelerating Reinforcement Learning of Robotic Manipulations via Feedback from Large Language Models},
  author={Chu, Kun and Zhao, Xufeng and Weber, Cornelius and Li, Mengdi and Wermter, Stefan},
  journal={arXiv preprint arXiv:2311.02379},
  year={2023}
}

@inproceedings{mahmoudieh2022zero,
  title={Zero-shot reward specification via grounded natural language},
  author={Mahmoudieh, Parsa and Pathak, Deepak and Darrell, Trevor},
  booktitle={International Conference on Machine Learning},
  pages={14743--14752},
  year={2022},
  organization={PMLR}
}

@inproceedings{radford2021learning,
  title={Learning transferable visual models from natural language supervision},
  author={Radford, Alec and Kim, Jong Wook and Hallacy, Chris and Ramesh, Aditya and Goh, Gabriel and Agarwal, Sandhini and Sastry, Girish and Askell, Amanda and Mishkin, Pamela and Clark, Jack and others},
  booktitle={International Conference on Machine Learning},
  pages={8748--8763},
  year={2021},
  organization={PMLR}
}

@InProceedings{ma2023liv,
  title = 	 {{LIV}: Language-Image Representations and Rewards for Robotic Control},
  author =       {Ma, Yecheng Jason and Kumar, Vikash and Zhang, Amy and Bastani, Osbert and Jayaraman, Dinesh},
  booktitle = 	 {Proceedings of the 40th International Conference on Machine Learning},
  pages = 	 {23301--23320},
  year = 	 {2023},
  editor = 	 {Krause, Andreas and Brunskill, Emma and Cho, Kyunghyun and Engelhardt, Barbara and Sabato, Sivan and Scarlett, Jonathan},
  volume = 	 {202},
  series = 	 {Proceedings of Machine Learning Research},
  month = 	 {23--29 Jul},
  publisher =    {PMLR},
  url = 	 {https://proceedings.mlr.press/v202/ma23b.html},
}

@inproceedings{sontakke2023roboclip,
author = {Sontakke, Sumedh A and Zhang, Jesse and Arnold, S\'{e}bastien M. R. and Pertsch, Karl and B\i{}y\i{}k, Erdem and Sadigh, Dorsa and Finn, Chelsea and Itti, Laurent},
title = {RoboCLIP: one demonstration is enough to learn robot policies},
year = {2023},
publisher = {Curran Associates Inc.},
address = {Red Hook, NY, USA},
booktitle = {Proceedings of the 37th International Conference on Neural Information Processing Systems},
articleno = {2430},
numpages = {13},
location = {New Orleans, LA, USA},
series = {NIPS '23}
}

@inproceedings{rocamonde2023visionlanguage,
title={Vision-Language Models are Zero-Shot Reward Models for Reinforcement Learning},
author={Juan Rocamonde and Victoriano Montesinos and Elvis Nava and Ethan Perez and David Lindner},
booktitle={NeurIPS 2023 Foundation Models for Decision Making Workshop},
year={2023},
url={https://openreview.net/forum?id=JUwczEJY8I}
}

@misc{adeniji2023language,
      title={Language Reward Modulation for Pretraining Reinforcement Learning}, 
      author={Ademi Adeniji and Amber Xie and Carmelo Sferrazza and Younggyo Seo and Stephen James and Pieter Abbeel},
      year={2023},
      eprint={2308.12270},
      archivePrefix={arXiv},
      primaryClass={cs.LG}
}

@InProceedings{cui2022zeroshot,
  title = {Can Foundation Models Perform Zero-Shot Task Specification For Robot Manipulation?},
  author = {Cui, Yuchen and Niekum, Scott and Gupta, Abhinav and Kumar, Vikash and Rajeswaran, Aravind},
  booktitle = {Proceedings of The 4th Annual Learning for Dynamics and Control Conference},
  pages = {893--905},
  year = {2022},
  editor = {Firoozi, Roya and Mehr, Negar and Yel, Esen and Antonova, Rika and Bohg, Jeannette and Schwager, Mac and Kochenderfer, Mykel},
  volume = {168},
  series = {Proceedings of Machine Learning Research},
  month = {23--24 Jun},
  publisher = {PMLR},
  url = {https://proceedings.mlr.press/v168/cui22a.html}
}

@inproceedings{ng1999invariance,
  author = {Ng, Andrew Y. and Harada, Daishi and Russell, Stuart J.},
  title = {Policy Invariance Under Reward Transformations: Theory and Application to Reward Shaping},
  year = {1999},
  isbn = {1558606122},
  publisher = {Morgan Kaufmann Publishers Inc.},
  address = {San Francisco, CA, USA},
  booktitle = {Proceedings of the Sixteenth International Conference on Machine Learning},
  pages = {278–287},
  numpages = {10},
  series = {ICML '99}
}

@inproceedings{devlin2012dynamicPBRS,
  author       = {Sam Devlin and
                  Daniel Kudenko},
  title        = {Dynamic potential-based reward shaping},
  booktitle    = {International Conference on Autonomous Agents and Multiagent Systems,
                  {AAMAS} 2012, Valencia, Spain, June 4-8, 2012 {(3} Volumes)},
  pages        = {433--440},
  publisher    = {{IFAAMAS}},
  year         = {2012},
  url          = {http://dl.acm.org/citation.cfm?id=2343638},
  bibsource    = {dblp computer science bibliography, https://dblp.org}
}

@inproceedings{grzes2017episodicPBRS,
author = {Grze\'{s}, Marek},
title = {Reward Shaping in Episodic Reinforcement Learning},
year = {2017},
publisher = {International Foundation for Autonomous Agents and Multiagent Systems},
address = {Richland, SC},
booktitle = {Proceedings of the 16th Conference on Autonomous Agents and MultiAgent Systems},
pages = {565–573},
numpages = {9},
location = {S\~{a}o Paulo, Brazil},
series = {AAMAS '17}
}

@article{mueller2025incorrectincomplete,
author={M{\"u}ller, Henrik
and Berg, Lukas
and Kudenko, Daniel},
title={Using incomplete and incorrect plans to shape reinforcement learning in long-sequence sparse-reward tasks},
journal={Neural Computing and Applications},
year={2025},
month={Jan},
day={10},
issn={1433-3058},
doi={10.1007/s00521-024-10615-2},
url={https://doi.org/10.1007/s00521-024-10615-2}
}

@inproceedings{mueller2025effectivepbrs,
author = {M\"{u}ller, Henrik and Kudenko, Daniel},
title = {Improving the Effectiveness of Potential-based Reward Shaping in Reinforcement Learning},
year = {2025},
isbn = {9798400714269},
publisher = {International Foundation for Autonomous Agents and Multiagent Systems},
address = {Richland, SC},
booktitle = {Proceedings of the 24th International Conference on Autonomous Agents and Multiagent Systems},
pages = {2684–2686},
numpages = {3},
location = {Detroit, MI, USA},
series = {AAMAS '25}
}

@article{Hasanbeig2021DeepSynthAS, 
    title={DeepSynth: Automata Synthesis for Automatic Task Segmentation in Deep Reinforcement Learning}, 
    volume={35}, 
    url={https://ojs.aaai.org/index.php/AAAI/article/view/16935}, 
    DOI={10.1609/aaai.v35i9.16935}, 
    number={9}, 
    journal={Proceedings of the AAAI Conference on Artificial Intelligence}, 
    author={Hasanbeig, Mohammadhosein and Yogananda Jeppu, Natasha and Abate, Alessandro and Melham, Tom and Kroening, Daniel}, 
    year={2021}, 
    month={May}, 
    pages={7647-7656}
}

@article{elbarbari2022tlrl,
author = {Elbarbari, Mahmoud and Delgrange, Florent and Vervlimmeren, Ivo and Efthymiadis, Kyriakos and Vanderborght, Bram and Nowe, Ann},
year = {2022},
month = {06},
pages = {},
title = {A framework for flexibly guiding learning agents},
journal = {Neural Computing and Applications},
doi = {10.1007/s00521-022-07396-x}
}

@inproceedings{suay2016demonstration,
author = {Suay, Halit Bener and Brys, Tim and Taylor, Matthew E. and Chernova, Sonia},
title = {Learning from Demonstration for Shaping through Inverse Reinforcement Learning},
year = {2016},
isbn = {9781450342391},
publisher = {International Foundation for Autonomous Agents and Multiagent Systems},
address = {Richland, SC},
booktitle = {Proceedings of the 2016 International Conference on Autonomous Agents \& Multiagent Systems},
pages = {429–437},
numpages = {9},
location = {Singapore, Singapore},
series = {AAMAS '16}
}

@inproceedings{wu2021demonstrations,
author = {Wu, Yuchen and Mozifian, Melissa and Shkurti, Florian},
title = {Shaping Rewards for Reinforcement Learning with Imperfect Demonstrations Using Generative Models},
year = {2021},
publisher = {IEEE Press},
url = {https://doi.org/10.1109/ICRA48506.2021.9561333},
doi = {10.1109/ICRA48506.2021.9561333},
booktitle = {2021 IEEE International Conference on Robotics and Automation (ICRA)},
pages = {6628–6634},
numpages = {7},
location = {Xi'an, China}
}

@inproceedings{brys2015demonstration,
author = {Brys, Tim and Harutyunyan, Anna and Suay, Halit Bener and Chernova, Sonia and Taylor, Matthew E. and Now\'{e}, Ann},
title = {Reinforcement Learning from Demonstration through Shaping},
year = {2015},
isbn = {9781577357384},
publisher = {AAAI Press},
booktitle = {Proceedings of the 24th International Conference on Artificial Intelligence},
pages = {3352–3358},
numpages = {7},
location = {Buenos Aires, Argentina},
series = {IJCAI'15}
}

@inproceedings{wang2023dshape,
author = {Wang, Caroline and Warnell, Garrett and Stone, Peter},
title = {D-Shape: Demonstration-Shaped Reinforcement Learning via Goal-Conditioning},
year = {2023},
isbn = {9781450394321},
publisher = {International Foundation for Autonomous Agents and Multiagent Systems},
address = {Richland, SC},
booktitle = {Proceedings of the 2023 International Conference on Autonomous Agents and Multiagent Systems},
pages = {1267–1275},
numpages = {9},
location = {London, United Kingdom},
series = {AAMAS '23}
}

@ARTICLE{zhang2025llmbackground,
  author={Zhang, Fuxiang and Li, Junyou and Li, Yi-Chen and Zhang, Zongzhang and Yu, Yang and Ye, Deheng},
  journal={IEEE Transactions on Neural Networks and Learning Systems}, 
  title={Improving Sample Efficiency of Reinforcement Learning With Background Knowledge From Large Language Models}, 
  year={2025},
  volume={},
  number={},
  pages={1-12},
  doi={10.1109/TNNLS.2025.3590731}}

\end{document}